# GCond: Gradient Conflict Resolution via Accumulation-Based Stabilization for Large-Scale Multi-Task Learning

**EVGENY ALVES LIMARENKO[1], ANASTASIIA STUDENIKINA[1,2], SVETLANA ILLARIONOVA[2,3], AND MAXIM SHARAEV[2,3]**
[1]Moscow Institute of Physics and Technology, 141701 Dolgoprudny, Russia
[2]Skolkovo Institute of Science and Technology, 121205 Moscow, Russia
[3]Biomedically Informed Artificial Intelligence Laboratory (BIMAI-Lab), University of Sharjah, Sharjah, United Arab Emirates

Corresponding author: Svetlana Illarionova (s.illarionova@skoltech.ru)

This work was supported by Russian Science Foundation under Grant 25-71-10088.

**ABSTRACT** In multi-task learning (MTL), gradient conflict poses a significant challenge. Effective methods for addressing this problem, including PCGrad, CAGrad, and GradNorm, in their original implementations are computationally demanding, which significantly limits their application in modern large models such as transformers. We propose Gradient Conductor (GCond), a method that builds upon PCGrad principles by combining them with gradient accumulation and an adaptive arbitration mechanism. We evaluated GCond on self-supervised multi-task learning tasks using MobileNetV3-Small and ConvNeXt architectures on the ImageNet 1K dataset and a combined head and neck CT scan dataset, comparing the proposed method against baseline linear combinations and state-of-the-art gradient conflict resolution methods. The classical and stochastic approaches of GCond were analyzed. The stochastic mode of GCond achieved a two-fold computational speedup while maintaining optimization quality, and demonstrated superior performance across all evaluated metrics, achieving lower L1 and SSIM losses compared to other methods on both datasets, and demonstrating superior generalization in heterogeneous scenarios: GCond improved ImageNet Top-1 Accuracy by 4.5% over baselines and prevented confidence overfitting in medical diagnosis tasks. GCond exhibited high scalability, being successfully applied to both compact models: MobileNetV3-Small and ConvNeXt-tiny; and large architecture ConvNeXtV2-Base. It also showed compatibility with modern optimizers such as AdamW and Lion/LARS. Therefore, GCond offers a scalable and efficient solution to the problem of gradient conflicts in multi-task learning.

**INDEX TERMS** Self-supervised learning, multi-task learning, gradient accumulation, gradient conflicts, PCGrad, medical imaging.

## I. INTRODUCTION

Multitask learning (MTL) enables neural networks to simultaneously optimize multiple related objective functions, which enhances generalization and computational efficiency compared to training separate models [1]. MTL can be viewed as a multi-criteria optimization problem that seeks to find an optimal solution in the space of trade-offs between multiple competing objectives under resource constraints [2]. It is important to distinguish between two types of multi-objective learning: classical multi-task learning with multiple tasks, such as simultaneous image classification and segmentation, and multi-objective optimization of a single task using a combination of multiple loss functions. The latter approach can be considered as a special case of multi-criteria optimization, where a single task is optimized through multiple quality metrics, for example, a combination of L1 and Structural Similarity Index (SSIM) for image reconstruction [3], [4]. One of the key factors influencing optimization dynamics is batch size. It determines the number of dataset elements processed in one iteration of the training process and has a substantial impact on model performance [5]. Modern research demonstrates that increasing batch size reduces model training time. In particular, a study by Jia and colleagues showed the possibility of reducing ResNet-50 training time on ImageNet from 29 hours to several minutes when increasing batch size from 256 to 64K [6]. Furthermore, the combination of high image resolution with large batch size demonstrates an average increase of 22% in accuracy for image classification and segmentation models [7]. However, practical application of large batches is limited by available GPU memory. Modern complex deep learning models require substantial amounts of memory for parameter storage, which significantly restricts the batch sizes that can be accommodated in device memory [8]. Exceeding the available memory necessitates reducing image resolution or decreasing batch size, which negatively affects final model quality [7]. Gradient accumulation methods offer an elegant solution to the limited memory problem while preserving the efficiency of large batches. This approach divides a large batch into K smaller micro-batches, which are processed sequentially with accumulation of computed gradients without exceeding the device memory limit [9]. After accumulating gradients across all microbatches, they are averaged to approximate the full batch gradient and the optimizer performs parameter updates. Mathematically, the gradient obtained through gradient accumulation is equivalent to the gradient computed for a large batch, ensuring the method's theoretical validity [10]. Although gradient accumulation increases training time due to replacing parallel computations with sequential ones, it is widely used in MTL tasks as it helps reduce the variance of the accumulated gradients [11]. A fundamental challenge in practical MTL implementation is accompanied by the fundamental problem of gradient conflicts, when gradients from different loss functions are directed in opposite directions, which manifests as negative cosine similarity between task gradients, and leads to mutual suppression of parameter updates [12]. This is accompanied by slower convergence, loss of important details, and suboptimal solutions. This effect is particularly pronounced in early training stages, when model parameters have not yet adapted to the multi-objective optimization landscape, which can cause loss of important details and deterioration of solution quality [13]. Experiments have confirmed that direct averaging of conflicting gradients leads to gradient bias, which can significantly worsen performance of individual tasks, since parameter updates become biased toward dominant tasks [14], [15], [16], [17]. The problem is exacerbated by the "tragic triad," when conflicting gradient directions combine with significant differences in magnitude and high curvature in the optimization landscape [18]. Under such conditions, traditional approaches demonstrate critical performance degradation compared to separate task training [19]. Modern approaches to gradient conflict resolution fall into three categories: naive summation [20], task balancing methods such as GradNorm [21], and projection methods such as Projecting Conflicting Gradients (PCGrad) [12] and Conflict-Alleviated Gradient Descent (CAGrad) [21]. Naive summation methods are based on joint task training with averaged loss functions and often with equal weights [20]. However, as shown in research, direct gradient averaging often worsens results compared to separate task training [18]. Task balancing approaches such as GradNorm dynamically adjust loss function weights to equalize their gradient magnitudes, but do not solve the direction problem, since under this approach even gradients with identical norms can be directed in opposite directions, creating conflict [23]. The most promising are considered to be projection methods or Gradient Surgery methods, among which PCGrad has become the de facto standard. PCGrad addresses conflicts by iterating through pairs of task gradients. For each conflicting pair, it projects one gradient onto the orthogonal complement of the other. The selection of which gradient to project is determined by their fixed order in the iteration, the effects of which are mitigated by randomly shuffling the task order at each training step [12]. CAGrad developed this idea, proposing the search for a common update direction by modifying gradients, transforming obtuse angles between them into acute ones and minimizing the average loss function [22]. PCGrad and CAGrad have been successfully used repeatedly in various computer vision tasks, including several medical imaging tasks [24], [25]. These methods have several significant limitations: they work on noisy gradients from single mini-batches and require substantial computational resources. Moreover, in their original implementation, PCGrad and CAGrad require significant memory costs due to the need to preserve computational graphs of weights between iterations, making them inapplicable in their original implementation for modern architectures like Transformer or ConvNeXt with large effective batch sizes [26]. Another approach is Gradient Sign Dropout (GradDrop), which solves the problem of gradient conflicts by randomly dropping elements of the gradients of the task based on element-wise sign mismatch (conflict) during backpropagation of the error [14]. Shi's research shows that GradDrop improves performance compared to simple averaging and can compete with more complex methods such as PCGrad or CAGrad on standard MTL benchmarks [19]. More modern projection methods for gradient conflict resolution such as Aligned-MTL, using object-level gradients instead of parameter-level gradients, or Similarity-Aware Momentum Gradient Surgery, dynamically adapting the gradient descent optimization process based on task gradient magnitude similarity, significantly reduce computational costs and accelerate training [27], [28]. However, to demonstrate their effectiveness, these methods rely on synthetic data, where artificial MTL tasks with controlled optimization conditions are used [29]. Such an approach may prove unviable when working with real medical data, characterized by high dimensionality, noise, data incompleteness, and complex nonlinear dependencies between tasks. In this work, we propose Gradient Conductor (GCond) - a method that builds upon the ideas of PCGrad by combining them with gradient accumulation. Unlike reactive strategies that eliminate conflicts at each step, GCond implements a two-phase "accumulate-then-resolve" process: during the accumulation phase, averaged gradients with low variance are computed. Then during the arbitration phase, an adaptive conflict resolution mechanism is applied, significantly improving the precision of gradient adjustments while reducing computational requirements. The contribution of our work can be described as follows:

1. We apply a novel approach to gradient conflict resolution based on accumulation, to reduce gradient variance before conflict resolution, which enhances stability and reliability of corrections.
2. We introduce an adaptive multi-zone arbitration mechanism with continuous conflict resolution strategies, which are based on learning dynamics (stability, strength, domination), for decision making.
3. Our approach demonstrates high scalability, performance, and efficiency on both single-channel medical imaging data and color natural images.

The underlying code for model training is shared: https://github.com/AlevLab-dev/GCond.

## II. MATH MATERIALS AND METHODS

### A. ACCUMULATION AS VARIANCE REDUCTION

The proposed GCond method organically integrates its gradient conflict resolution mechanism-projecting conflicting gradients onto each other, similar to PCGrad-with gradient accumulation. Specifically, instead of operating on "noisy" gradients g($\theta$;b) from individual mini-batches, which are unbiased but highvariance estimates of the true gradient G($\theta$), GCond utilizes accumulated gradients. GCond addresses this problem through a two-phase approach: accumulation and arbitration. In the first phase (Estimation Phase), over K steps, gradients for each of the N loss functions are computed on different mini-batches and accumulated in N independent accumulator buffers. This is followed by the Resolution Phase: after K steps, the proposed adaptive arbitrator mechanism is applied to the N accumulated, averaged gradients to resolve conflicts and produce a single, unified gradient, which is then passed to the optimizer. GCond uses gradient accumulation to obtain averaged gradients over K steps:

$$\hat{\boldsymbol{g}}_i = \frac{1}{K}\sum_{k=1}^{K}\boldsymbol{g}_i\left(\theta; b_k\right) \tag{1}$$

where $\hat{\mathbf{g}}_\mathbf{i}$ is the accumulated gradient for task i, K is the number of accumulation steps.

Consequently, the variance of the accumulated gradients is significantly reduced:

$$\mathrm{Var}\left(\hat{\boldsymbol{g}}_i\right) = \frac{1}{K}\,\mathrm{Var}\left(\boldsymbol{g}_i\right) \tag{2}$$

This provides more stable and reliable estimates of the true gradient directions, enabling precise detection and resolution of conflicts (assuming independent or weakly correlated micro-batches). Thus, GCond employs a standard gradient accumulation mechanism to suppress noise, which fundamentally distinguishes it from other methods. This transforms the conflict resolution procedure into a more robust analysis of the true descent directions of the loss functions. Such an approach turns metrics of inter-gradient interactions, like the cosine similarity between accumulated gradients $\hat{\mathbf{g}}_\mathbf{i}$ and $\hat{\mathbf{g}}_\mathbf{j}$, from high-variance random variables into statistically reliable indicators of the true conflict between tasks. This ensures more informed and stable gradient correction decisions.

### B. ADAPTIVE MULTI-ZONE ARBITRATION

After obtaining the accumulated gradients $\{\hat{\mathbf{g}}_\mathbf{1}, \ldots, \hat{\mathbf{g}}_\mathbf{n}\}$, our arbitration mechanism resolves conflicts iteratively, starting with the most severe ones. At each iteration, the pair of gradients ($\hat{\mathbf{g}}_\mathbf{i}$, $\hat{\mathbf{g}}_\mathbf{j}$) with the minimum cosine similarity is identified:

$$c = \cos\left(\hat{\boldsymbol{g}}_i, \hat{\boldsymbol{g}}_j\right) = \frac{\hat{\boldsymbol{g}}_i \cdot \hat{\boldsymbol{g}}_j}{\left\|\hat{\boldsymbol{g}}_i\right\|\left\|\hat{\boldsymbol{g}}_j\right\|} \tag{3}$$

Instead of discretely switching between resolution strategies, our approach employs a continuous modulation governed by a piecewise function. This function non-linearly maps the cosine similarity c $\in [-1,1]$ to an effective conflict angle $\alpha$eff $\in [0, \pi]$, guided by three thresholds: $\theta_{\text{weak}}$, $\theta_{\text{main}}$, and $\theta_{\text{crit}}$.

$$\alpha_{\text{eff}}(c) = \begin{cases} \pi, & \text{if } c < \theta_{\text{crit}} \\ \frac{\pi}{2} + \frac{\pi}{2}\left(\frac{c-\theta_{\text{main}}}{\theta_{\text{crit}}-\theta_{\text{main}}}\right)^{p}, & \text{if } \theta_{\text{crit}} \leq c < \theta_{\text{main}} \\ \frac{\pi}{2}\left(1 - \frac{c-\theta_{\text{main}}}{\theta_{\text{weak}}-\theta_{\text{main}}}\right), & \text{if } \theta_{\text{main}} \leq c < \theta_{\text{weak}} \end{cases} \tag{4}$$

Here, p is a hyperparameter (remap_power) controlling the mapping's curvature. The projection strengths for the winner ($g_w$) and loser ($g_l$) gradients are then determined by two trigonometric modulators derived from this angle: a winner scaling factor

$$s_w = \sin(\alpha_{\text{eff}}) \tag{5}$$

and a loser scaling factor

$$s_l = \sin(\min\{\alpha_{\text{eff}}, \pi/2\}) \tag{6}$$

This mechanism ensures smooth, continuous transitions between conflict resolution behaviors, which are detailed in Table 1. Conflicts where c $\geq \theta_{\text{weak}}$ are considered agreement, and no correction is applied.

The method for identifying the "winner" and "loser" in conflicting pairs is described in the following section.

### C. WINNER SELECTION

In situations requiring arbitration (i.e., in moderate and critical conflict zones where c $< \theta_{\text{weak}}$), we identify a "winner"

**TABLE 1. Zonal classification of gradient conflicts.**

| Conflict Zone | Condition | Resolution Strategy |
|---|---|---|
| Agreement | $c \geq \theta_{weak}$ | No correction. The arbitration loop terminates. |
| Mild Conflict | $\theta_{main} \leq c < \theta_{weak}$ | Symmetric Scaled Projection. $\alpha_{eff}$ maps to $(0,\pi/2]$. Both scaling factors, $s_w$ and $s_l$, decay from 1 to 0. Projections are applied symmetrically, with their magnitude diminishing as $c$ approaches $\theta_{weak}$. |
| Moderate Conflict | $\theta_{crit} \leq c < \theta_{main}$ | Asymmetric Projection. $\alpha_{eff}$ maps to $(\pi/2,\pi]$. The loser is fully projected ($s_l = 1$). The winner receives a partial corrective projection that starts at zero and strengthens as $c$ approaches $\theta_{main}$ ($s_w$ grows from 0 to 1). |
| Critical Conflict | $c < \theta_{crit}$ | Winner preservation. $\alpha_{eff}$ is fixed at $\pi$. The winner's gradient is unchanged ($s_w = 0$), while the loser is fully projected onto the winner's orthogonal complement ($s_l = 1$). |

gradient using a weighted score:

$$\text{Score}_i = w_{\text{stability}} \cdot \max(0, S_i) + w_{\text{strength}} \cdot N_i \quad (7)$$

where: Si (Stability): Represents the cosine similarity between the current accumulated gradient and the gradient from the previous optimization step:

$$S_i = \cos\left(\hat{\mathbf{g}}_i(t), \hat{\mathbf{g}}_i(t-1)\right) \quad (8)$$

Stability rewards gradients that maintain their direction, leading to a smoother training trajectory. $N_i$ (Strength): A relative measure of a gradient's magnitude. It is calculated by first normalizing the gradient's norm by its Exponential Moving Average (EMA), and then scaling this value by the sum of the scaled norms for both conflicting gradients:

$$N_i = \frac{\|\hat{g}_i\|_2 / \left(\text{EMA}\left(\|\hat{g}_i\|_2\right) + \varepsilon\right)}{\sum_{k=i,j} \|\hat{g}_k\|_2 / \left(\text{EMA}\left(\|\hat{g}_k\|_2\right) + \varepsilon\right)} \quad (9)$$

where strength prioritizes gradients that are momentarily stronger relative to their historical average and relative to their current competitor. This allows for a rapid response to tasks that suddenly become more important. Additionally, a dominance prevention mechanism is implemented: if the same task wins conflicts for a specified number of iterations (dominance window), its opponent is automatically designated the winner. This prevents the training of other tasks from stagnating. A schematic representation of the architectural overview of GCond is presented in Figures 1, and an adaptive multi-zone arbitration mechanism is described in Figures 9 of the Appendix.

### D. STOCHASTIC MODE

To enhance computational efficiency, especially with a large number of accumulation steps ($K \gg 1$), a stochastic mode for GCond was developed. In this mode, the K accumulation steps are divided into N non-overlapping blocks of K/N steps for each loss function. The gradient for each task i is accumulated over its unique sequence of mini-batches. It is important to note that all gradients $\mathbf{g_i}$ are computed with respect to the same model weight state $\theta_t$, ensuring that the final accumulated gradients $\hat{\mathbf{g}}_i$ are unbiased and comparable. This approach not only reduces computational costs, but also promotes more robust gradient estimates by analyzing data from different samples.

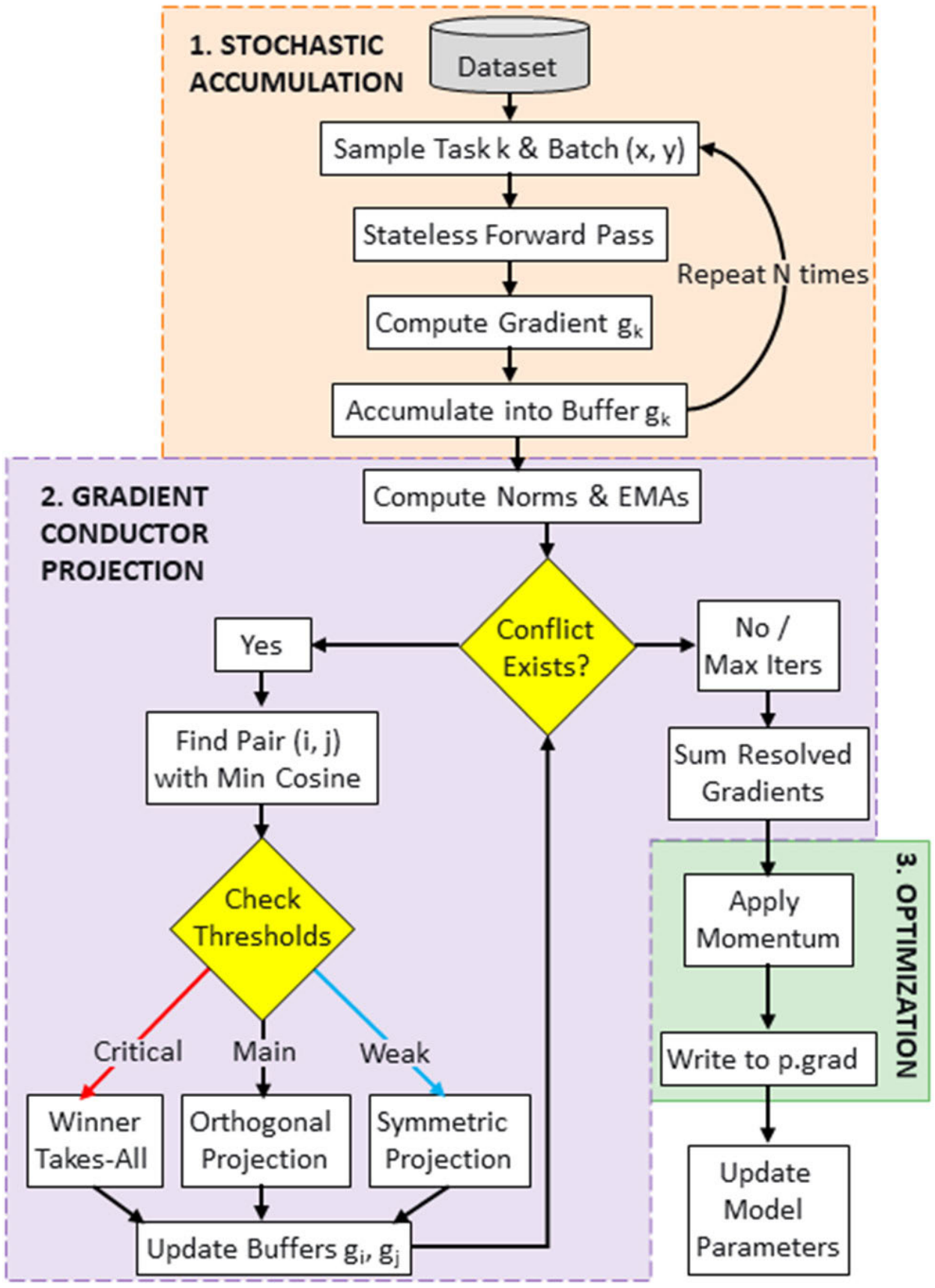

**FIGURE 1. Architectural overview of the GCond.**
**Note. The general structure of GCond consists of the following stages: stochastic accumulation - gradients from N tasks are calculated independently over K iterations; gradient conductor projection (resolution phase) - the adaptive arbitrator iteratively resolves conflicts between the accumulated gradients based on cosine similarity; conflicts are managed using a multi-zone strategy (critical, main, weak), before they are summed; finally, optimization occurs.**

### E. EXPERIMENTAL SETTINGS

#### 1) COMPUTATIONAL ENVIRONMENT

Experiments with simpler models MobileNetV3-Small and ConvNeXt-tiny were conducted on a system with an NVIDIA RTX 4080 GPU (16 GB VRAM) and a 20 core Intel(R) Core (TM) Ultra 7 265K CPU. The software environment included

PyTorch 2.7.0, Python 3.12.3, Ubuntu 24.04, and CUDA 12.8. Experiments with model ConvNeXtV2-Base were conducted on a system with an NVIDIA RTX 5090 GPU (32 GB VRAM) AMD EPYC 9654 96-Core Processor PyTorch 2.5.1, Python 3.10.18, Ubuntu 22.04, and CUDA 12.4.

#### 2) EXPERIMENTAL DATASETS

To ensure fair and reproducible comparisons, all experiments were conducted under strictly controlled conditions. We implemented a self-supervised image reconstruction task (Masked Image Modeling) on a large dataset of medical head and neck CT scans from three public datasets: RSNA Intracranial Hemorrhage Detection [30], RSNA Cervical Spine Fracture Detection [31], and RADCURE from The Cancer Imaging Archive (TCIA) [32], containing 2,199,444 DICOM (Digital Imaging and Communications in Medicine) files. All experiments used a fixed patient-based data split to ensure reproducibility and fair method comparison. Additionally, the performance of the proposed GCond model was verified on benjamin-paine/imagenet-1k-256 $\times$ 256-a version of the ILSVRC 2012 (ImageNet) dataset containing 1.28 million training images and 50,000 validation images.

#### 3) EXPERIMENTAL CONFIGURATION

In our experiments, we used a Masked Autoencoder (MAE) architecture with a MobileNetV3-Small encoder and a 2−layer Transformer decoder (embedding dimension 256). For scalability analysis, a ConvNeXt-tiny architecture was used.

#### 4) TRAINING CONFIGURATION

All models were trained for 15 epochs with a linear warmup for 2 epochs, followed by a cosine learning rate decay. We used the AdamW optimizer with a learning rate of $2.0 \times 10^{-4}$ and a weight decay of 0.05. A large effective batch size of $256 \times 24 = 6144$ was used for training stability and to reduce gradient variance. Input images were preprocessed by center cropping and resizing to a resolution of $256 \times 256$ pixels.

#### 5) LOSS FUNCTIONS

Two loss functions, calculated on the masked patches, were used: L1 loss to enforce pixel-level accuracy ($\lambda_{L1} = 0.85$) and SSIM loss to promote structural similarity ($\lambda_{SSIM} = 0.15$). We employed the L1 metric paired with SSIM, because L1 is more robust to the outliers and noise characteristic of medical CT data, leading to more stable training than L2. The primary focus of this study is the mechanism for resolving gradient conflicts, not the optimization of loss function design. A key aspect of our experimental design was the deliberate separation of the value spaces for these two metrics. The L1 loss was calculated directly in the z-normalized space, whereas the SSIM loss was computed in the standard pixel-value range of [0, 1] after applying an inverse normalization to both the model's predictions and the target patches. This approach was chosen specifically to exacerbate the natural conflict between the pixelwise L1 metric and the perceptual SSIM metric, thereby creating a more demanding test scenario for evaluating the effectiveness of gradient conflict resolution strategies.

#### 6) EVALUATION INDICATORS

Model quality was assessed using L1 Loss and SSIM Loss on the validation set. Efficiency was evaluated based on peak VRAM consumption (MB) and throughput (samples/sec). Stability was monitored through gradient norms and loss curve dynamics. Robustness was assessed using Top-1, Top-5 and Cross-Entropy metrics for ImageNet, as well as AUC-ROC and BCE metrics for RSNA, using the ConvNeXtV2-Base architecture.

#### 7) BENCHMARKS

The proposed GCond algorithm was benchmarked against a standard weighted summation baseline and state-of-the-art (SOTA) gradient projection methods: PCGrad, CAGrad, GradDrop, and GradNorm, implemented from their respective papers.

### F. HYPERPARAMETER SELECTION STRATEGY

The hyperparameters of GCond are derived from geometric principles of optimization and validated through ablation studies detailed in Appendix. We categorize them into three groups: physical constants, geometric thresholds, and arbitration policies

#### 1) OPTIMIZATION CONSTANTS

to ensure a continuous differentiable response to conflict intensity, avoiding the optimization instability associated with discrete switching logic, use_smooth_logic = True and remap_power = 2.0 are recommended as default. This configuration introduces a quadratic non-linearity into the _get_effective_alpha function, ensuring smoother gradient modulation in the weak conflict zone ($\cos(\theta) \to \theta_{weak}$) and more decisive correction as the conflict intensifies (as $\cos(\theta) \to \theta_{crit}$), thereby stabilizing training dynamics. The momentum_beta parameter should follow standard adaptive optimizer values (e.g., 0.9) to mitigate stochastic noise.

#### 2) GEOMETRIC THRESHOLDS

Intervention zones are defined based on the cosine similarity ($\cos\theta$) between gradients. The default thresholds tau $\theta_{crit} = -0.8$ (strong opposition), $\theta_{main} = -0.5$ (obtuse angle), and $\theta_{weak} = 0.0$ (orthogonality) are grounded in Euclidean geometry. While $\theta_{weak} = 0.0$ is the theoretical baseline for independence, a transient warmup phase shifting $\theta_{weak}$ from 0.1 to −0.1 may be beneficial in early training to enforce initial directional consensus.

#### 3) ARBITRATION POLICY

To resolve deadlocks and prevent "chattering" (oscillation between competing gradients), an asymmetric tie-breaking strategy is required (e.g., 0.8/0.2), favoring either stability or relative strength to ensure a consistent update direction.

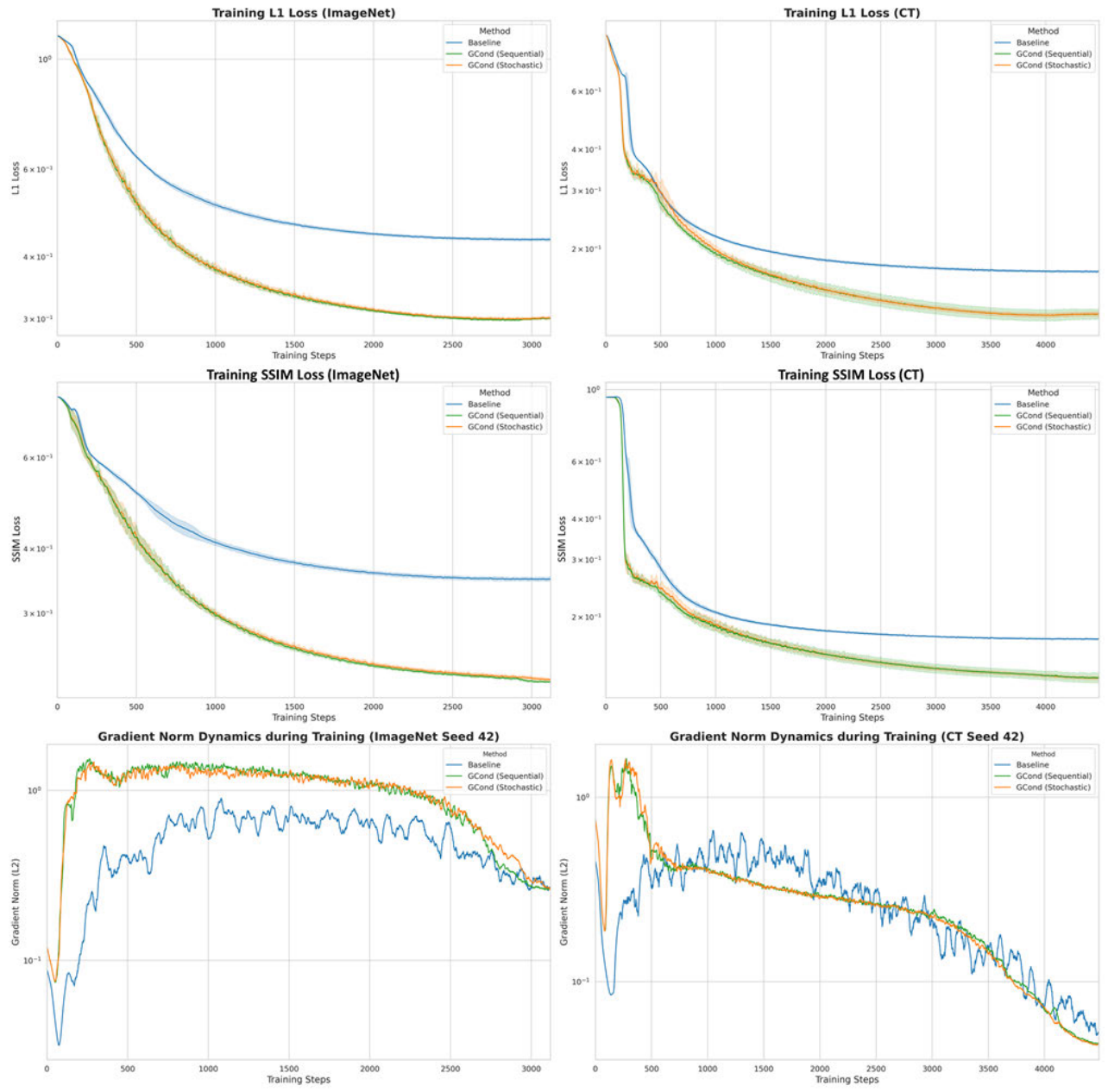

**FIGURE 2. Comparison of convergence for L1 and SSIM losses, and L2-norms of gradients for the stochastic and exact GCond modes on ImageNet and CT datasets during MobileNetV3-Small model training. Note. x-axis: Training steps, y-axis (top row): L1 Loss, y-axis (middle row): SSIM Loss, y-axis (bottom row): Gradient Norm.**

This physically based approach has allowed us to identify the most reliable configuration, which provides an effective balance between stability and convergence speed, and requires minimal tuning for standard multi-task scenarios. For more information on the hyperparameter tables, please see the Appendix section.

#### 4) USE OF ARTIFICIAL INTELLIGENCE

The manuscript has been edited for English language consistency and grammatical accuracy using generative artificial intelligence tools, in particular Google's Gemini. The authors reviewed and edited the text and bear full responsibility for the final content.

## III. RESULTS

### A. PERFORMANCE ANALYSIS OF THE STOCHASTIC GCOND MODE

A comparison of the training dynamics between the stochastic and the exact (sequential) modes is presented in Fig. 2. The results demonstrate that the differences between the learning curves are statistically insignificant. This confirms the hypothesis that with a sufficiently large effective batch size (in this case, $256 \times 24 = 6144$), a sparse sampling of gradients allows for the formation of a statistically reliable estimate for conflict resolution. Notably, the stochastic mode exhibited a twofold increase in computational efficiency compared to the sequential mode, while preserving training quality. The validation curves confirm these observations (Appendix, Fig. 10).

### B. COMPARATIVE ANALYSIS OF TRAINING PROCESS DYNAMICS AND OPTIMIZATION QUALITY

To evaluate the effectiveness of the proposed GCond method, a comparison was conducted against SOTA gradient conflict resolution approaches and a baseline linear combination of loss functions (Baseline).

Fig. 3 illustrates the dynamics of the loss functions and the L2-norms of gradients during training on the CT HN and ImageNet datasets, respectively. Across all loss plots, it is observed that simple task weighting (Baseline) yields more stable and smoother training trajectories compared to traditional gradient conflict resolution methods. On both datasets, SOTA methods consistently underperformed the Baseline in terms of L1 loss. Given the large effective batch size (6144), the Baseline method benefits from inherent gradient smoothing, exhibiting low variance and high stability. Consequently, the Baseline proves more reliable than SOTA methods. These SOTA approaches, designed for "noisy" mini-batches, apply aggressive corrections and "hard" projections at every step, making them less suitable for our low-variance environment.

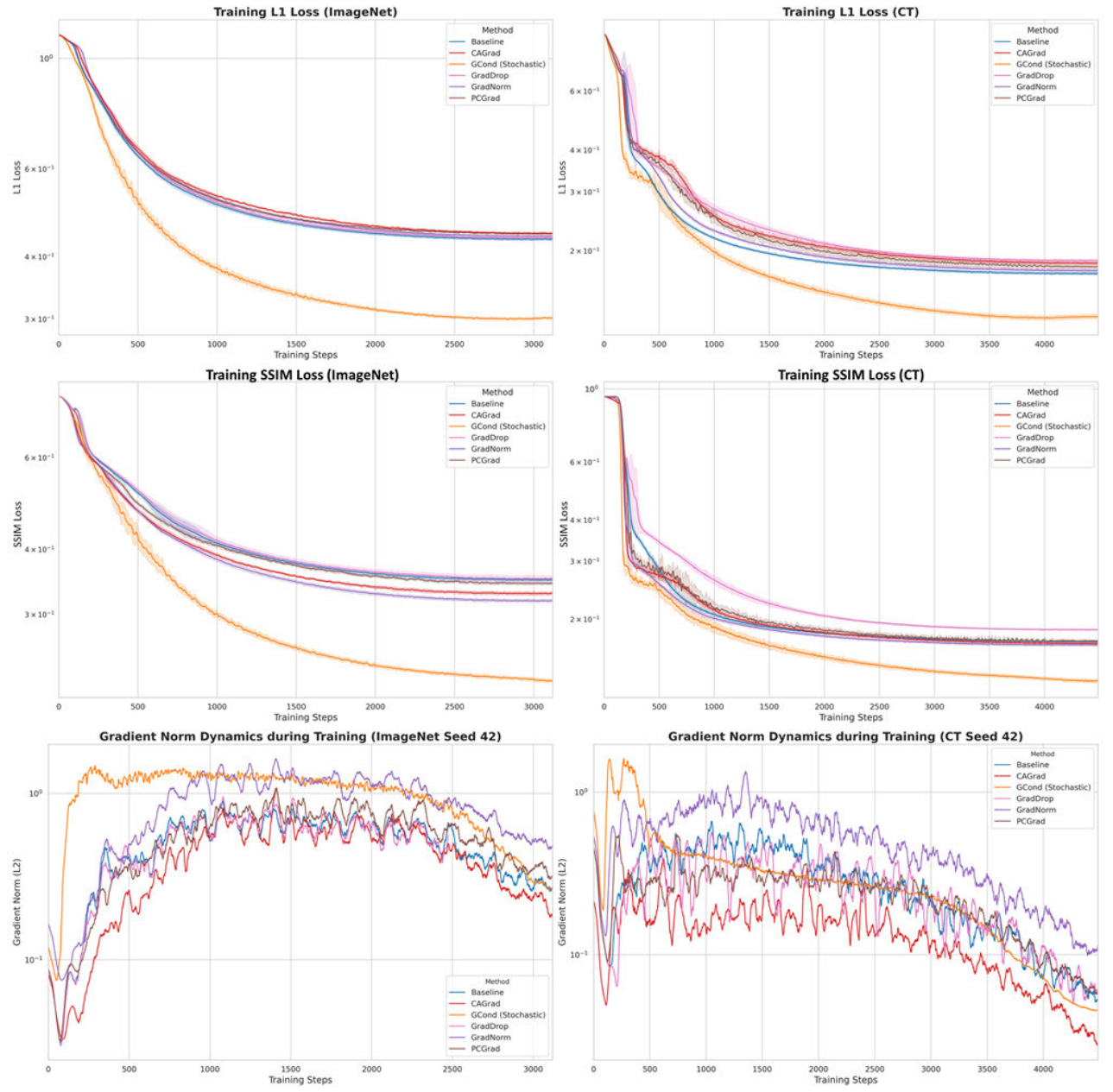

**FIGURE 3. Comparison of the dynamics of L1 and SSIM loss functions, and L2-norms of gradients for all methods on both datasets during MobileNetV3-Small model training.**
**Note. x-axis: Training steps, y-axis (top row): L1 Loss, y-axis (middle row): SSIM Loss, y-axis (bottom row): Gradient Norm.**

Regarding the SSIM loss on the ImageNet dataset, although SOTA methods showed better results than the Baseline, their training curves were accompanied by noticeable high-frequency oscillations and earlier stagnation on a plateau. In contrast, the stochastic mode of GCond demonstrates a clear superiority over all compared methods. The dynamics of the L1 and SSIM loss functions during training directly correspond to the dynamics of the gradient L2-norms. It is evident that GradNorm, GradDrop and PCGrad produce significant sawtooth-like oscillations in amplitude, while CAGrad systematically suppresses the gradient norm, especially on the CT HN dataset, which is equivalent to an excessively small effective learning step. The reasons for this lie in the nature of these algorithms, as they primarily focus on handling mini-batch gradients that are often sharply anisotropic. Against this backdrop, the simple summed gradient of the Baseline remains an unbiased, low-variance estimate of the descent direction and thus proves to be more reliable. The proposed GCond mechanism exhibits more monotonic and less noisy behavior because destructive projections are mitigated smoothly, and the step direction is statistically justified and largely invariant to the scale of the layers. By aggregating gradients over a large effective batch and utilizing an arbitration mechanism, GCond forms a globally consistent update vector. Its normalization and EMA-smoothing stabilize the step magnitude, allowing the Adam optimizer to perform larger and more reliable updates in the chosen direction. These conclusions are further illustrated by the validation curves (Appendix, Fig. 11).

The results demonstrate the consistent superiority of GCond over both the Baseline and other conflict resolution methods across both datasets (Table 2). Notably, the projection-based methods, PCGrad, GradDrop and CAGrad, performed worse than the Baseline. A key finding is the statistical equivalence between the sequential and stochastic GCond modes. Their mean performance metrics are nearly identical, with significantly overlapping confidence intervals suggesting no statistically significant difference This finding provides robust justification for utilizing the more computationally efficient stochastic mode in all subsequent experiments.

**TABLE 2. Best validation metrics for the MobileNetV3-Small model.**

| Method | ImageNet | | CT HN | |
|---|---|---|---|---|
| | L1 Loss | SSIM Loss | L1 Loss | SSIM Loss |
| Baseline | 0.41542 ± 0.00716 | 0.34845 ± 0.00764 | 0.16473 ± 0.00307 | 0.16145 ± 0.00154 |
| CAGrad | 0.42263 ± 0.00323 | 0.32793 ± 0.00643 | 0.17779 ± 0.00485 | 0.16096 ± 0.00222 |
| GradDrop | 0.42036 ± 0.01572 | 0.35241 ± 0.01571 | 0.17909 ± 0.00411 | 0.17218 ± 0.00336 |
| GradNorm | 0.41635 ± 0.00204 | 0.31818 ± 0.00579 | 0.16847 ± 0.00496 | 0.15930 ± 0.00241 |
| PCGrad | 0.42324 ± 0.00270 | 0.34221 ± 0.00462 | 0.17311 ± 0.00683 | 0.16286 ± 0.00371 |
| GCond (Sequential) | 0.31493 ± 0.00056 | 0.24485 ± 0.00093 | 0.12942 ± 0.01195 | 0.13118 ± 0.01049 |
| GCond (Stochastic) | 0.31655 ± 0.00287 | 0.24704 ± 0.00262 | 0.12941 ± 0.00405 | 0.13111 ± 0.00315 |

## C. ANALYSIS OF COMPUTATIONAL EFFICIENCY

To ensure a fair comparison, the number of accumulation steps was fixed across all experiments, and for each implementation, only the batch size was increased to fill 16 GB of VRAM. This maintains an equal number of updates per epoch and allows for a correct comparison of method throughput. The GCond implementation is based on torch.func.functional_call with an explicit call to autograd.grad on "fresh" leaf parameters: the gradients for each task are computed independently, immediately accumulated into bf16 buffers, and the computation graph is released. This approach eliminates the need to retain a common graph until the last task (retain_graph=True in classic PCGrad/CAGrad/GradNorm), meaning GCond's peak memory consumption is close to that of a single backward pass plus a fixed overhead for the accumulation buffers. This explains the slightly higher VRAM usage on the small MobileNetV3-Small model Table 3.

**TABLE 3. Comparative performance of methods on the MobileNetV3-Small model.**

| Method | ImageNet | | | CT HN | | |
|---|---|---|---|---|---|---|
| | GPU Memory (MB) | Epoch Time (s) | Throughput (samples/s) | GPU Memory (MB) | Epoch Time (s) | Throughput (samples/s) |
| Baseline | 6888 | 901 | 1421 | 4257 | 778 | 2367 |
| CAGrad | 7514 | 965 | 1327 | 4682 | 989 | 1861 |
| GradNorm | 7056 | 965 | 1327 | 4319 | 993 | 1853 |
| PCGrad | 7515 | 968 | 1322 | 4679 | 993 | 1853 |
| GCond | 8739 | 905 | 1415 | 4856 | 822 | 2237 |

On the ConvNeXtV2-Base architecture with 16 GB of VRAM, the PCGrad and CAGrad methods failed to run even with a batch size of 1. This was due to the necessity of retaining the computation graph and individual gradient vectors, as the memory consumption of these methods scales linearly with the number of tasks. In contrast, GCond was able to process up to 70 images thanks to its graph-free task processing. Therefore, the ConvNeXt-tiny model was used to compare the speed of the algorithms Table 4. With a fixed number of accumulation steps, GCond achieves Baseline throughput and surpasses PCGrad/CAGrad in performance (throughput) with comparable memory consumption (VRAM). Thus, a key advantage of GCond is its scalability.

**TABLE 4. Comparative performance of methods on the ConvNeXt-tiny model, CT HN dataset Seed 42.**

| Method | Batch Size | Effective Batch Size | GPU Memory (MB) | Epoch Time (s) | Throughput (samples/s) |
|---|---|---|---|---|---|
| PCGrad | 96 | 2304 | 14193 | 4817 | 382 |
| CAGrad | 96 | 2304 | 14182 | 4872 | 378 |
| GradNorm | 148 | 3552 | 13959 | 4992. | 369 |
| Baseline | 170 | 4080 | 14511 | 3199 | 576 |
| GCond | 162 | 3888 | 14540 | 3205 | 574 |

### D. MULTI-TASK LEARNING

We extended the evaluation to a ConvNeXtV2-Base backbone across two distinct multi-task scenarios: RSNA Intracranial Hemorrhage detection and ImageNet-1k classification, both coupled with a dense masked autoencoder (MAE) reconstruction auxiliary task. Experiments reveal that GCond's impact adapts to data complexity.

**TABLE 5. Comparative evaluation of MTL strategies on RSNA intracranial hemorrhage with ConvNeXtV2-Base.**

| Method | L1 Loss | 1 - SSIM | Macro AUC | BCE Loss |
|---|---|---|---|---|
| Baseline | 0.0785 | 0.0615 | 0.9820 | 0.3961 |
| CAGrad | 0.1056 | 0.0814 | 0.9822 | 0.4068 |
| GCond | 0.0701 | 0.0567 | 0.9795 | 0.4062 |
| GradDrop | 0.0853 | 0.0675 | 0.9814 | 0.4058 |
| GradNorm | 0.0736 | 0.0528 | 0.9809 | 0.3921 |
| PCGrad | 0.0809 | 0.0581 | 0.9815 | 0.4098 |

**Note.** The table presents the best validation metrics observed during training

On the smaller RSNA dataset, GCond achieves the lowest L1 reconstruction loss (0.0701), significantly outperforming the Baseline and geometric projection methods (Table 5). GCond acted as a dynamic regularizer: while SOTA methods (PCGrad, GradNorm) exhibited severe confidence overfitting – characterized by diverging validation BCE losses (>1.0) despite high AUCs – GCond maintained calibration (Val BCE $\sim$ 0.55) by leveraging reconstruction gradients to ground semantic features.

**TABLE 6. Comparative evaluation of MTL strategies on imagenet-1k classification and reconstruction with ConvNeXtV2-Base.**

| Method | L1 Loss | 1 - SSIM | Top-1 Acc | Top-5 Acc | CE Loss |
|---|---|---|---|---|---|
| Baseline | 0.3151 | 0.2415 | 55.0800 | 78.0580 | 2.0682 |
| CAGrad | 0.3479 | 0.2579 | 56.0200 | 78.5700 | 2.0155 |
| GCond | 0.2677 | 0.1941 | 59.3240 | 81.8800 | 1.8793 |
| GradDrop | 0.3173 | 0.2425 | 54.6680 | 77.7820 | 2.0735 |
| GradNorm | 0.2888 | 0.1858 | 54.8740 | 78.0080 | 2.0577 |
| PCGrad | 0.3166 | 0.2400 | 55.3040 | 78.0840 | 2.0340 |

Conversely, on the large-scale ImageNet benchmark, GCond functioned as a plasticity facilitator, mitigating the early feature saturation observed in baselines. GCond demonstrates superior performance across all semantic metrics (Top-1/Top-5 Accuracy, Cross-Entropy Loss) and reconstruction objectives (Masked Patch L1, SSIM), effectively mitigating feature collapse observed in competing methods. This resulted in improvement in Top-1 Accuracy and a substantial mitigation of the U-shaped overfitting curve in classification loss (Appendix, Fig. 13). Across both domains, GCond consistently yielded the lowest reconstruction errors (L1/SSIM), demonstrating its superior capability in resolving conflicts between intensity-based and structural gradient signals compared to geometric projection or normalization strategies.

### E. INTEGRATION WITH ADAMW OPTIMIZERS

The interaction between GCond and the AdamW optimizer was investigated using two schemes: Integrated: GCond performs the projection and EMA-smoothing of the gradient, while AdamW is used only for RMS-normalization ($\beta_1 = 0$). Separate (Pure): GCond performs only the projection, with its internal EMA-smoothing disabled. AdamW operates in its standard mode ($\beta_1 > 0, \beta_2 > 0$). Results indicate a significant advantage for the Integrated Scheme (Fig. 4). This is due to the operational order: applying AdamW's adaptive denominator to an already harmonized and smoothed gradient vector ensures stability. In the separate scheme, the internal momentum of AdamW attempts to smooth an already-corrected but still non-stationary sequence of gradients, leading to an overestimated variance and premature training stagnation.

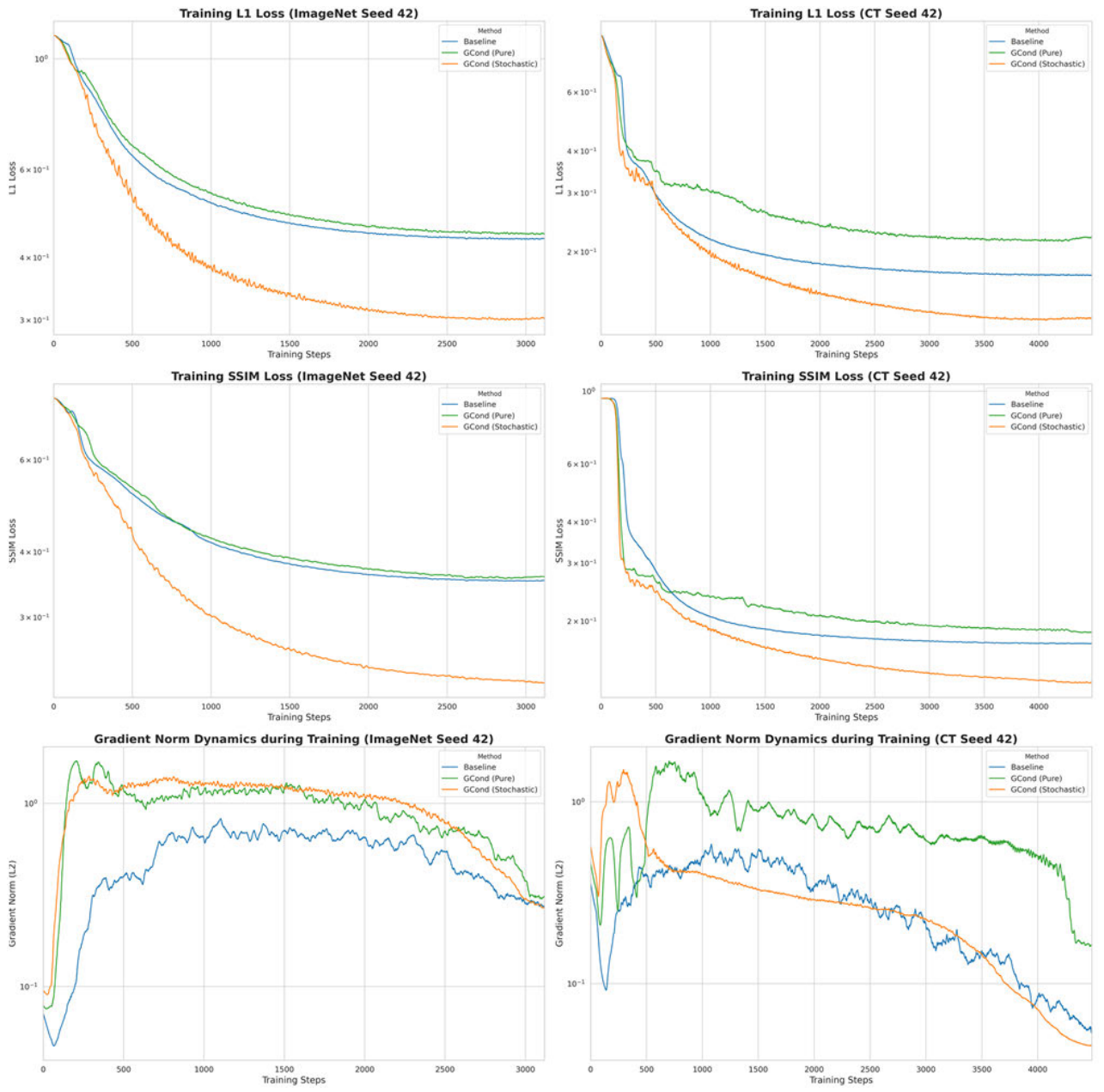

**FIGURE 4. Comparison of convergence for L1 and SSIM losses, and L2-norms of gradients for the stochastic and pure GCond modes, and Baseline on both datasets during MobileNetV3-Small model training. Note. x-axis: Training steps, y-axis (top row): L1 Loss, y-axis (middle row): SSIM Loss, y-axis (bottom row): Gradient Norm.**

### F. EXPERIMENTS WITH THE LION/LARS OPTIMIZER

The integrated GCond mode was tested with a hybrid optimizer combining principles from Lion and LARS: the update direction is given by sign(mt), where mt is the smoothed momentum, and the magnitude is scaled by an adaptive learning rate (trust-ratio):

$$\frac{\|\mathbf{p}\|}{\|\mathbf{m_t}\|} \tag{10}$$

(where p are the model weights). This approach combines the decisiveness of Lion, which ignores gradient magnitude

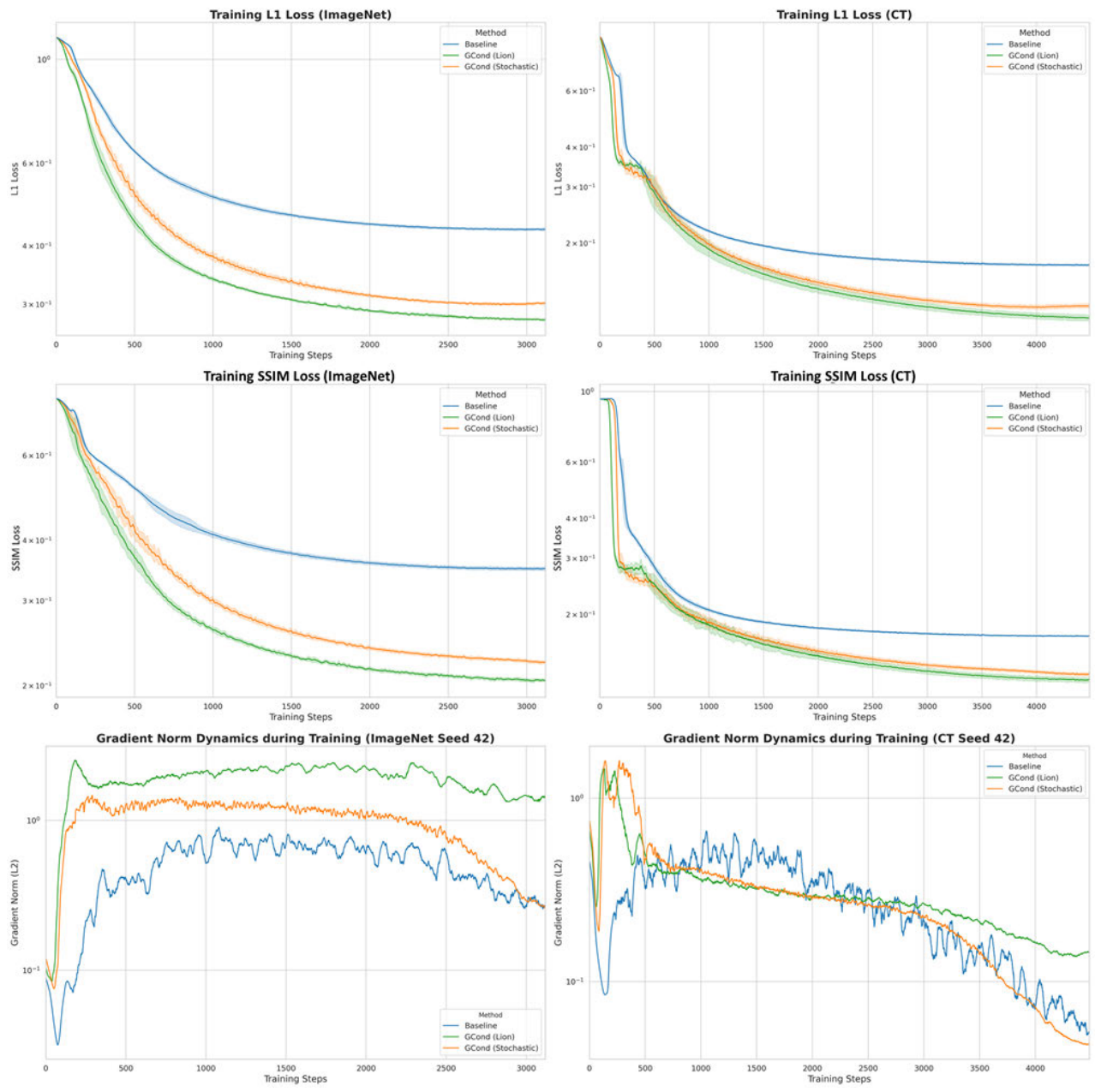

**FIGURE 5. Comparison of convergence for L1 and SSIM losses, and L2-norms for the stochastic GCond mode and GCond with the Lion/LARS optimizer, and Baseline on both datasets during MobileNetV3-Small model training.**
**Note. x-axis: Training steps, y-axis (top row): L1 Loss, y-axis (middle row): SSIM Loss, y-axis (bottom row): Gradient Norm.**

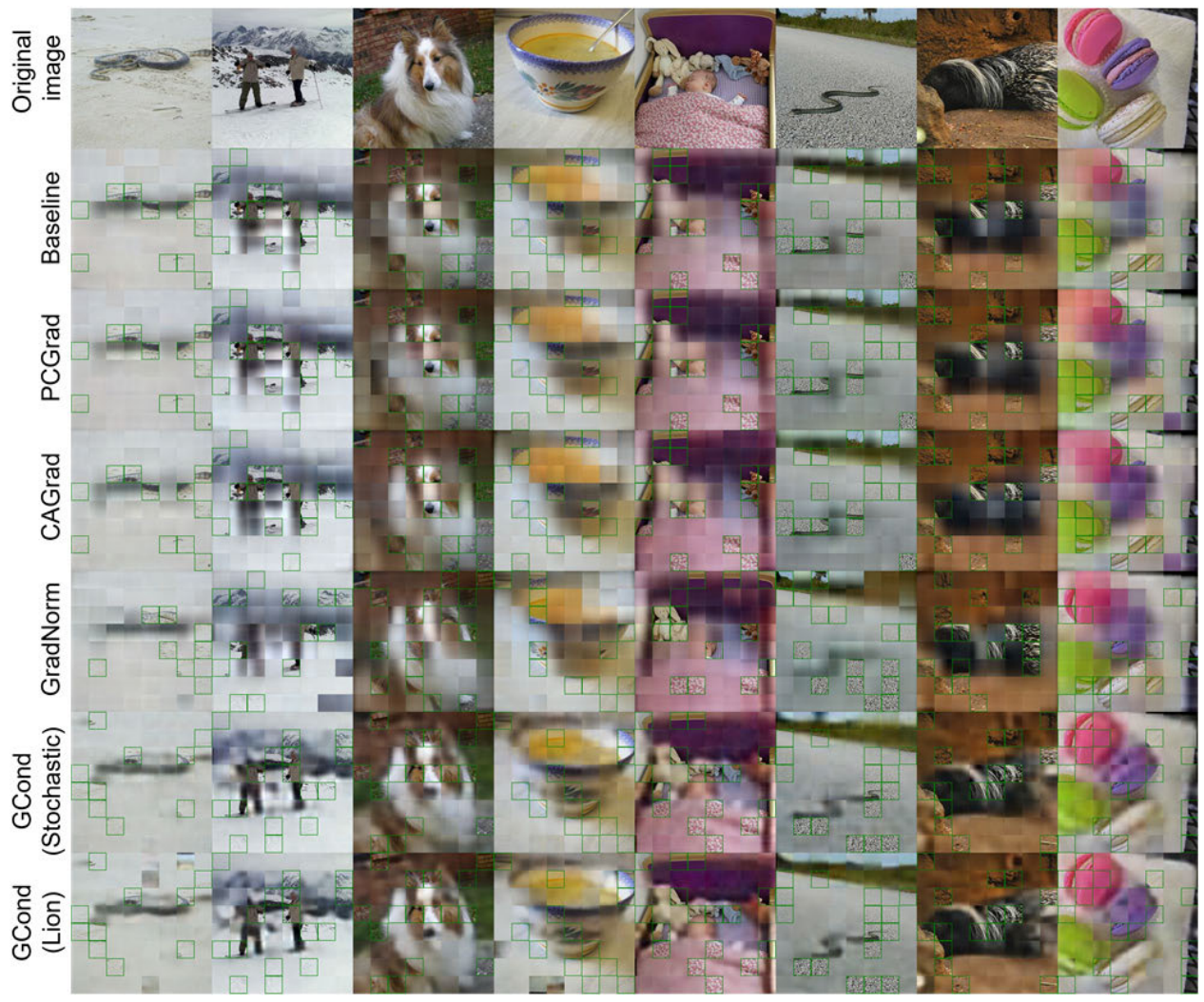

**FIGURE 6. Comparison of image reconstructions. The cells highlighted in green were visible to the model.**

to prevent getting stuck in narrow local minima, with the stabilizing normalization of LARS.

This resulted in a more confident loss reduction and convergence to a lower plateau compared to AdamW (Fig. 5) demonstrates that the Lion/LARS-GCond configuration achieved a substantial reduction in L1 and SSIM losses compared to the AdamW-GCond baseline. Furthermore, replacing AdamW with Lion reduces VRAM consumption by eliminating the need to store second-order moment estimates, which is critical for training large-scale architectures. As shown by the validation curves in (Appendix, Fig. 12).

It should be emphasized that a systematic comparison of all analyzed methods with the Lion and LARS optimizers was intentionally not conducted, and this result should be considered a promising direction for future research.

### G. VISUAL ANALYSIS OF RECONSTRUCTION QUALITY

Qualitative analysis on the ImageNet validation set (Fig. 6) confirms that the stochastic GCond mode, particularly when coupled with the Lion/LARS optimizer, yields visually superior image reconstructions. The improvements are particularly noticeable in the restoration of fine details, textures, and the preservation of the overall image structure.

## IV. DISCUSSION

Our analysis reveals characteristic limitations inherent to existing gradient conflict resolution methods. GradNorm and PCGrad create significant sawtooth-like oscillations in amplitude, while CAGrad systematically over-suppresses the gradient norm. The reasons for this behavior lie in the nature of these algorithms. The core idea of GradNorm is to dynamically adjust loss weights to equalize the learning rates of different tasks by normalizing the norms of gradients passing through shared network layers, preventing one task from dominating others [21]. However, re-tuning GradNorm's weights based on the instantaneous norms of a single shared layer provides a noisy proxy signal, especially under mixed-precision (AMP) and accumulation settings, leading to excessive corrections. PCGrad iteratively manipulates gradients by projecting the gradient of one task onto the normal plane of another's when they conflict [12]. In doing so, PCGrad's hard projections can nullify useful gradient components during a conflict, stochastically reducing the effective optimization step size. CAGrad extends the idea of PCGrad, seeking a single update vector that minimizes conflict with all tasks simultaneously [22]. However, CAGrad's internal optimization problem often finds a convex combination with additional normalization, which reduces the gradient norm and can exacerbate underfitting. GradDrop algorithmically selects one sign based on gradient distribution at the current iteration, and then masks out all task gradient values that have the opposite sign [14]. Thus, despite the theoretical elegance of existing gradient conflict resolution methods, our experiments show that their continuous and unconditional application can be counterproductive.

In phases where gradients are nearly co-directional, applying these methods can remove useful shared components, slowing convergence. Furthermore, these methods do not consider the optimization history, making them vulnerable to shortterm gradient oscillations. Against this backdrop, the simple summed gradient of the Baseline remains an unbiased, lower-variance estimate of the descent direction and therefore proves to be a more reliable solution when using a large effective batch size.

Crucially, our supplementary analysis on diverse architectures reveals a dual failure mode in these methods. On large-scale classification (ImageNet), they suffer from feature collapse, underfitting semantic tasks due to the dominance of dense reconstruction gradients (resulting in a 4.5% accuracy gap). Conversely, on smaller medical datasets (RSNA), they permit confidence overfitting, where Validation BCE diverges (>1.0) despite high AUC, indicating poor probability calibration. In contrast, GCond acts as a dynamic plasticity regulator: it facilitates feature learning on large-scale data by leveraging dense signals to navigate sparse classification landscapes, while simultaneously regularizing training on smaller datasets to prevent semantic hallucinations. This adaptability is absent in rigid projection methods.

In contrast, our comparative analysis demonstrates the significant superiority of the proposed GCond method over existing conflict resolution techniques and the Baseline approach. This is because, unlike the aforementioned methods, GCond implements a multi-phase, adaptive strategy that does not apply corrections unnecessarily. It forms a single gradient with a high signal-to-noise ratio through a suite of complementary mechanisms. These include: the functional computation of individual task gradients while temporarily setting the model to eval() mode (disabling Dropout and using running statistics for BatchNorm); a smoothed projection with conflict angle remapping; a "winner-loser" arbitration mechanism based on stability relative to the previous step and EMAs of norms; and finally, EMA smoothing and bias correction before feeding the gradient to the optimizer. GCond's operation can be characterized by three distinct phases. The Initial Phase, where active projection correction is applied to quickly exit regions of strong conflict. This is followed by the Harmonious Descent Phase, where projections are automatically deactivated when gradients become co-directional ($\cos\theta \approx 1$), preventing excessive manipulation and accelerating learning. The final stage is the Plateau Navigation Phase. When learning slows, GCond detects the re-emergence of weak conflicts and reactivates its corrective mechanism, enabling it to find paths for further loss reduction where other methods stagnate. A key element of this approach is the hybrid arbitration, which considers not only the current geometry of the gradients but also their historical stability (similarity to the previous step) and relative strength when selecting a dominant task. This allows GCond to avoid unfavorable compromises and find an optimization trajectory that effectively improves both target metrics. As a result, GCond's trajectory exhibits more monotonic and less noisy behavior; destructive projections are mitigated smoothly, and the step direction is statistically grounded and virtually invariant to the scale of the layers.

### *A. ARCHITECTURAL ADVANTAGES*

The development of the stochastic GCond mode is a significant methodological contribution, as it solves the critical problem of computational scalability while preserving optimization quality. The fact that the differences between the stochastic and exact modes are within the range of statistical noise Fig. 2 and validation curves in (Appendix, Fig. 10) confirms the hypothesis that sparse gradient sampling is sufficient to form a statistically reliable estimate of conflicts when using large effective batch sizes. The primary advantage of GCond is its scalability on large models. We note that while any method could theoretically be rewritten using torch.func to partially reduce memory overhead, the superiority of our approach is not merely implementational. It stems from the symbiosis of gradient accumulation with a smooth, statistically-grounded conflict resolution logic. In contrast, an ablation study of a "GCond-as-PCGrad" variant (using hard projections) shows a noticeable degradation in quality and efficiency, although it still outperforms a naive PCGrad implementation with random projection order. While methods like PCGrad and CAGrad in their original implementations were unable to run on the ConvNeXtV2-Base architecture-as their computational complexity and memory requirements grow at least linearly with the number of tasks-GCond successfully processed up to 70 images per batch. This was achieved by sequentially computing gradients for specific tasks within dedicated accumulation windows without retaining the computation graph, making GCond an ideal solution for training large modern models and transformers under limited computational resources where memory is the bottleneck. Experiments with ConvNeXtV2-Base demonstrated not only quantitative improvements in metrics but also qualitative changes in model training. The nearly twofold expansion in the activation distribution of the final encoder layer from the very first epochs indicate the formation of richer, non-collapsing representations. We attribute this to the high signal-to-noise ratio maintained by the conflict-resilient projection, which potentially enhances the model's generalization capability. Furthermore, GCond integrates exceptionally well with modern optimizers. The integration with Lion [33], which determines the step direction via the sign operation, and LARS [34], which adaptively scales the step magnitude, demonstrated superior results compared to the standard stochastic GCond mode with AdamW. The combination of Lion's decisiveness, which ignores gradient magnitude, with the stabilizing normalization of LARS creates a synergistic effect, ensuring a faster and more stable reduction in the loss function. This decomposition of tasks-geometry handled separately from step adaptation-maintains moderate orthogonality between tasks throughout training, allowing for continuous fine-grained adjustments that lead to better final solutions. The comparison between the "pure" GCond variant (without internal smoothing) and the standard integrated version highlighted the importance of the order of operations in adaptive optimization. The integrated scheme, where AdamW's adaptive denominator operates on an already-agreed-upon and smoothed gradient vector, provides significantly more stable update steps. This allows AdamW (with $\beta_1$ disabled to avoid redundant momentum) to take large, consistent steps and sharply accelerates convergence. The even lower L1/SSIM losses achieved by replacing the optimizer with Lion confirm that the quality of the direction is more critical than its magnitude alone. Together, these factors explain the sharp reduction in L1 and SSIM for the GCond method on both datasets and the absence of the training trajectory noise characteristic of naive gradient conflict methods. This indicates the potential for further development of specialized optimizers for multi-task learning.

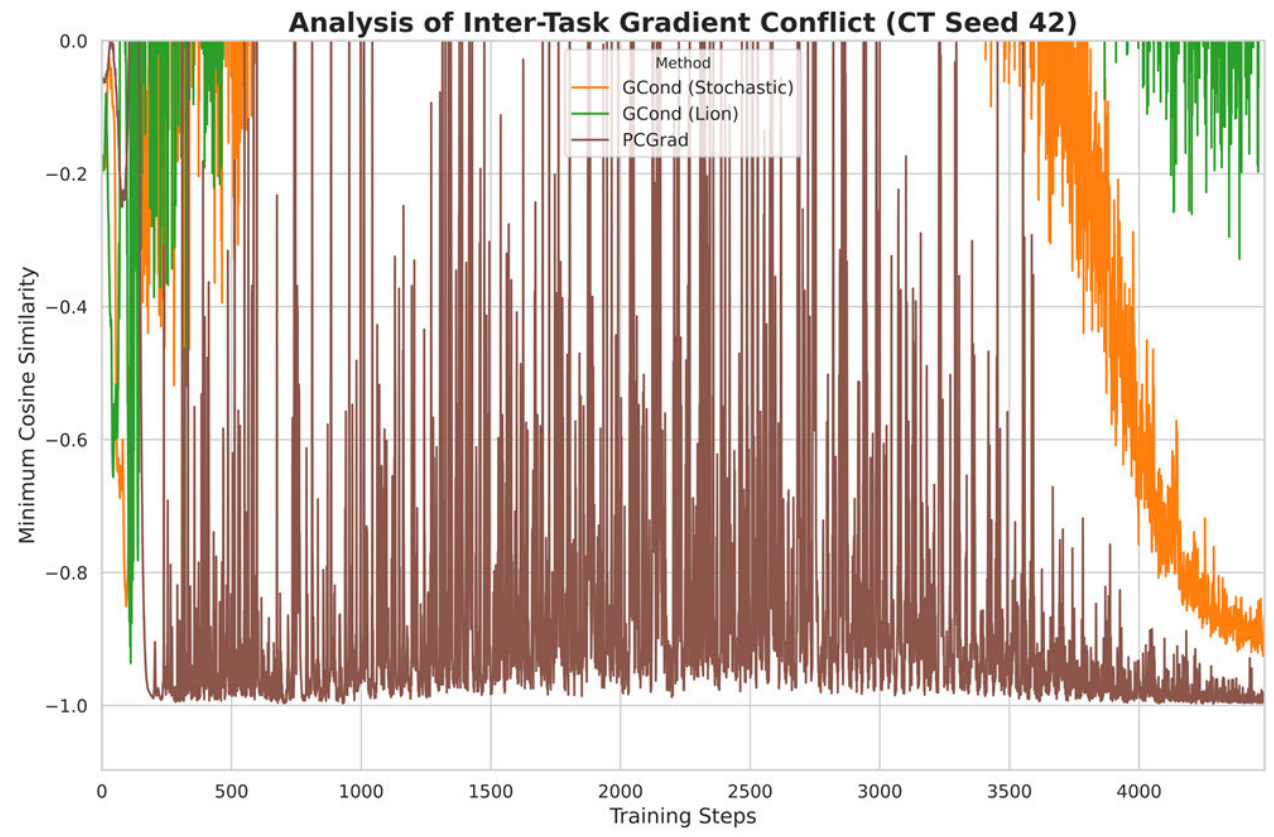

**FIGURE 7.** Cosine similarity of the most conflicting gradient pair at each step on the CT dataset.

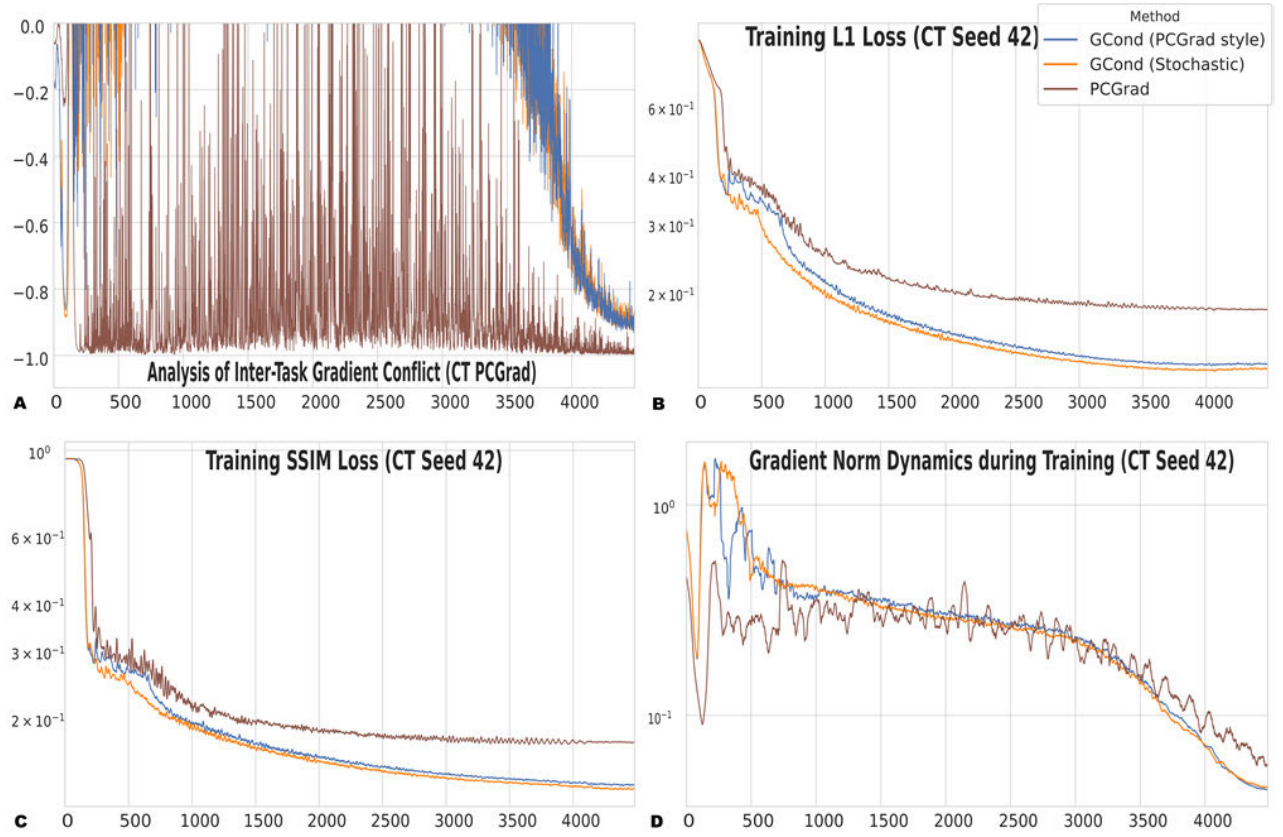

**FIGURE 8.** Cosine similarity of the most conflicting gradient pair at each step on the CT dataset and comparison of L1 and SSIM loss dynamics for PCGrad, "GCond-as-PCGrad," and stochastic GCond on the CT dataset. Note. x-axis: Training steps, y-axis: (A) Minimum c osine similarity; (B) L1 Loss; (C) SSIM Loss; (D) Gradient Norm.

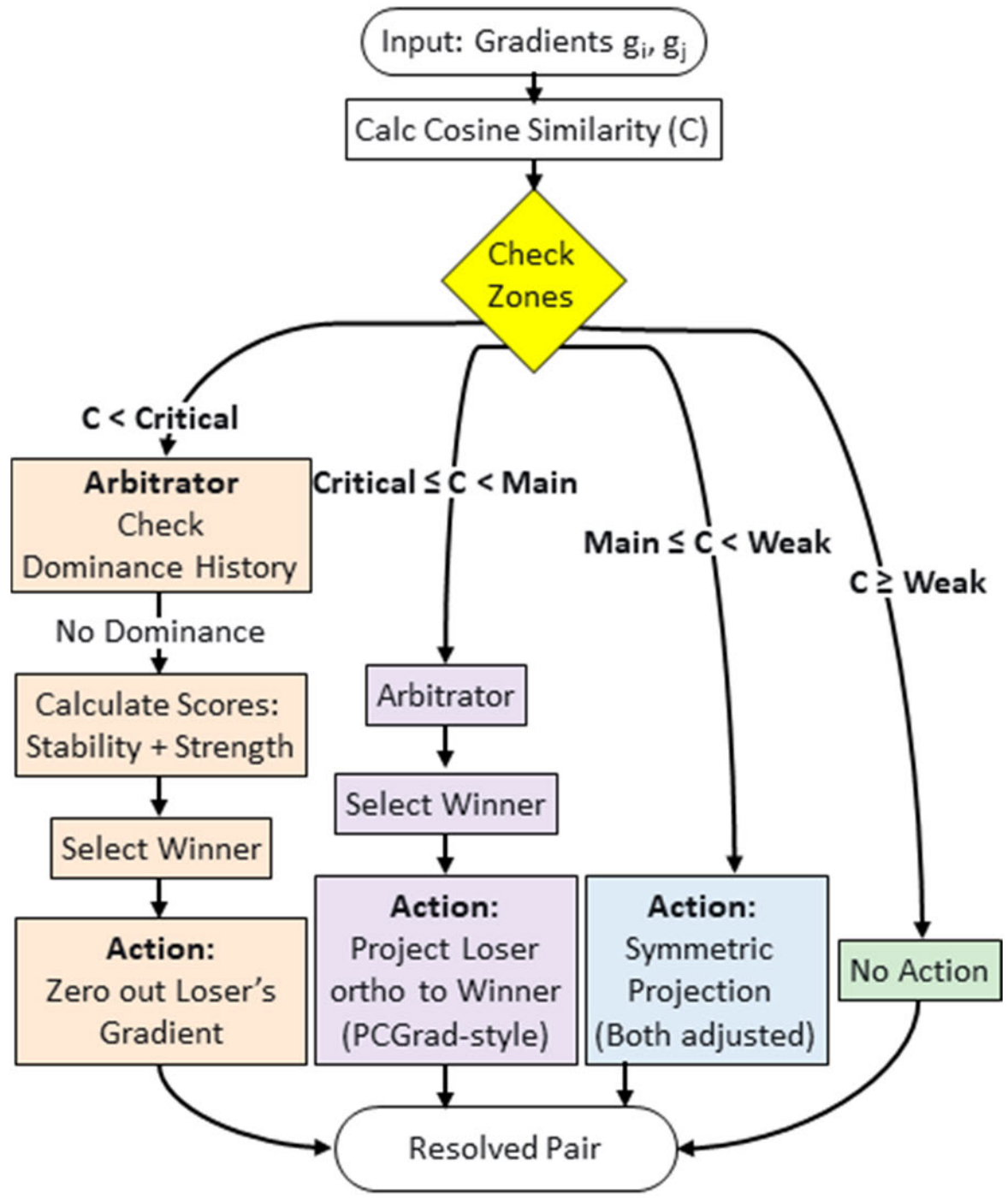

**FIGURE 9.** Flowchart of the adaptive multi-zone arbitration mechanism.

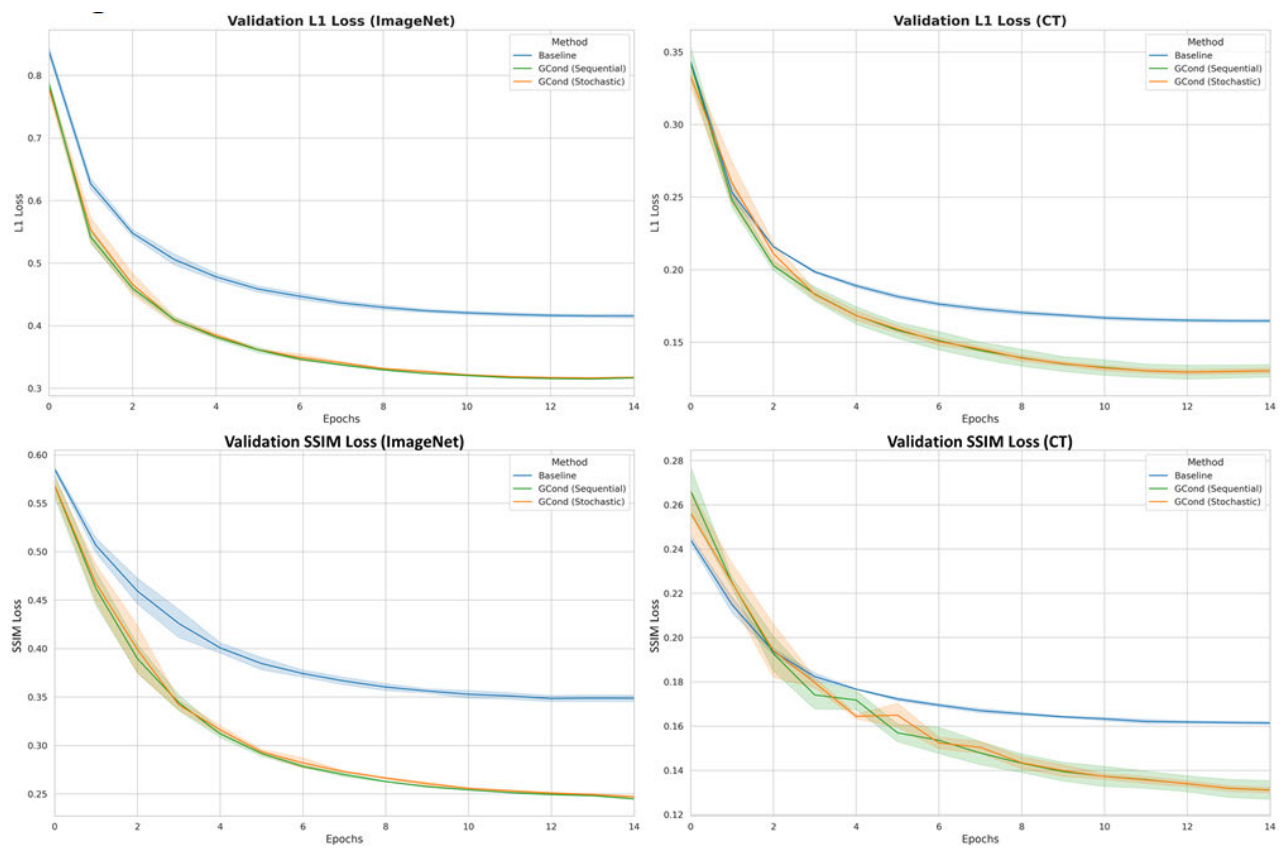

**FIGURE 10.** Validation of L1 and SSIM loss functions for the stochastic and exact GCond modes on both datasets during MobileNetV3-Small model training.
Note. x-axis: Epoch, y-axis (top row): L1 Loss, y-axis (bottom row): SSIM Loss.

## B. ANALYSIS OF CONFLICT ANGLE DYNAMICS

The curves of minimum cosine similarity show that GCond, particularly the Lion variant, quickly dampens early acute conflicts and transitions the multi-task dynamics into a stable "collinear" mode for most of the training trajectory (Fig. 7). Sporadic late-stage activations represent brief, targeted projections just before the model's final fine-tuning. The stochastic mode without Lion also remains passive until the later stages; however, as the capacity of MobileNetV3-Small is exhausted and filter specialization increases, an "active competition for parameters" emerges. The minimum cosine similarity drops, and the arbitrator engages a soft projection with angle remapping to locally redistribute the task contributions.

This stability is explained by GCond's architecture. First, the hierarchy of conflict zones with nonlinear angle remapping prevents hyper-correction during minor disagreements. Second, the arbitration based on "stability × strength" criteria with dominance memory suppresses oscillations and cycles. Third, the EMA-norms in the stochastic mode and the corrected momentum with LARS-tuning in the Lion version equalize the scale of the directions. For comparison, PCGrad, by constantly performing pairwise orthogonalization, keeps the minimum cosines in the negative region, locking the system in a conflict mode and destroying the agreed-upon descent direction. This explains why GCond does not intervene for most of the training process but resolves conflicts in a targeted manner during the final stages, which correlates with the continued reduction in L1 and SSIM losses after all other compared methods have plateaued.

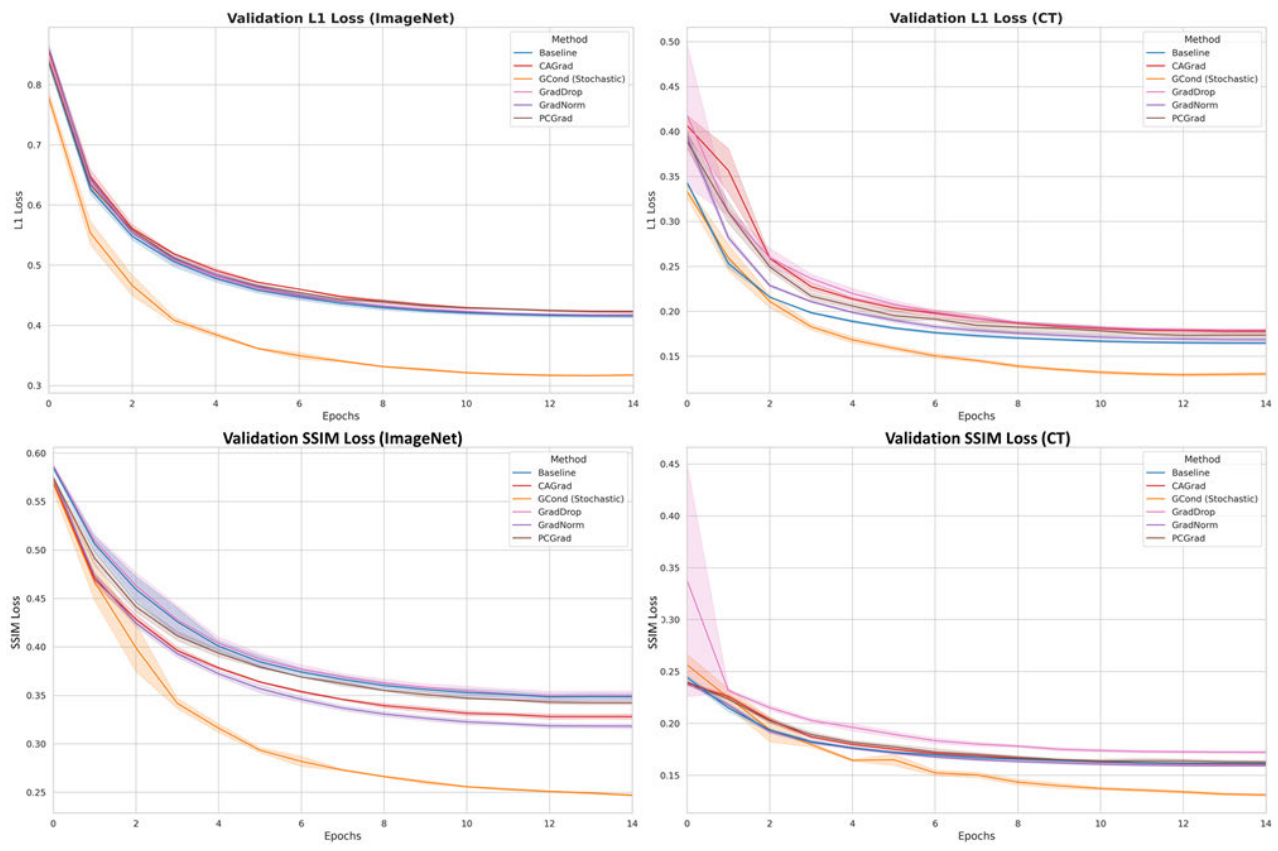

**FIGURE 11.** **Validation of L1 and SSIM loss functions for all methods on both datasets during MobileNetV3-Small model training.**
**Note. x-axis: Epoch, y-axis (top row): L1 Loss, y-axis (bottom row): SSIM Loss.**

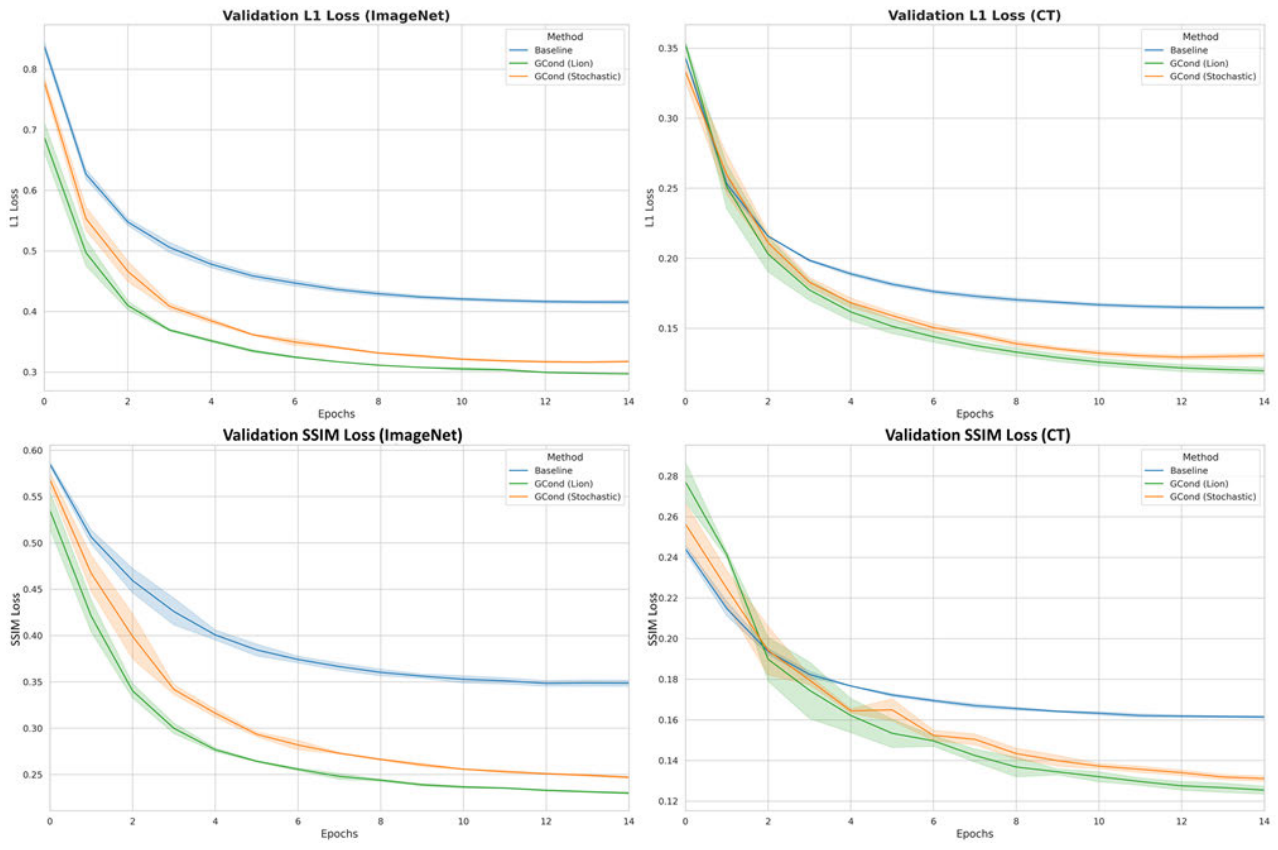

**FIGURE 12.** **Validation of L1 and SSIM loss functions for the stochastic GCond mode and GCond with the Lion/LARS optimizer, and Baseline on both datasets during MobileNetV3-Small model training.**
**Note. x-axis: Epoch, y-axis (top row): L1 Loss, y-axis (bottom row): SSIM Loss.**

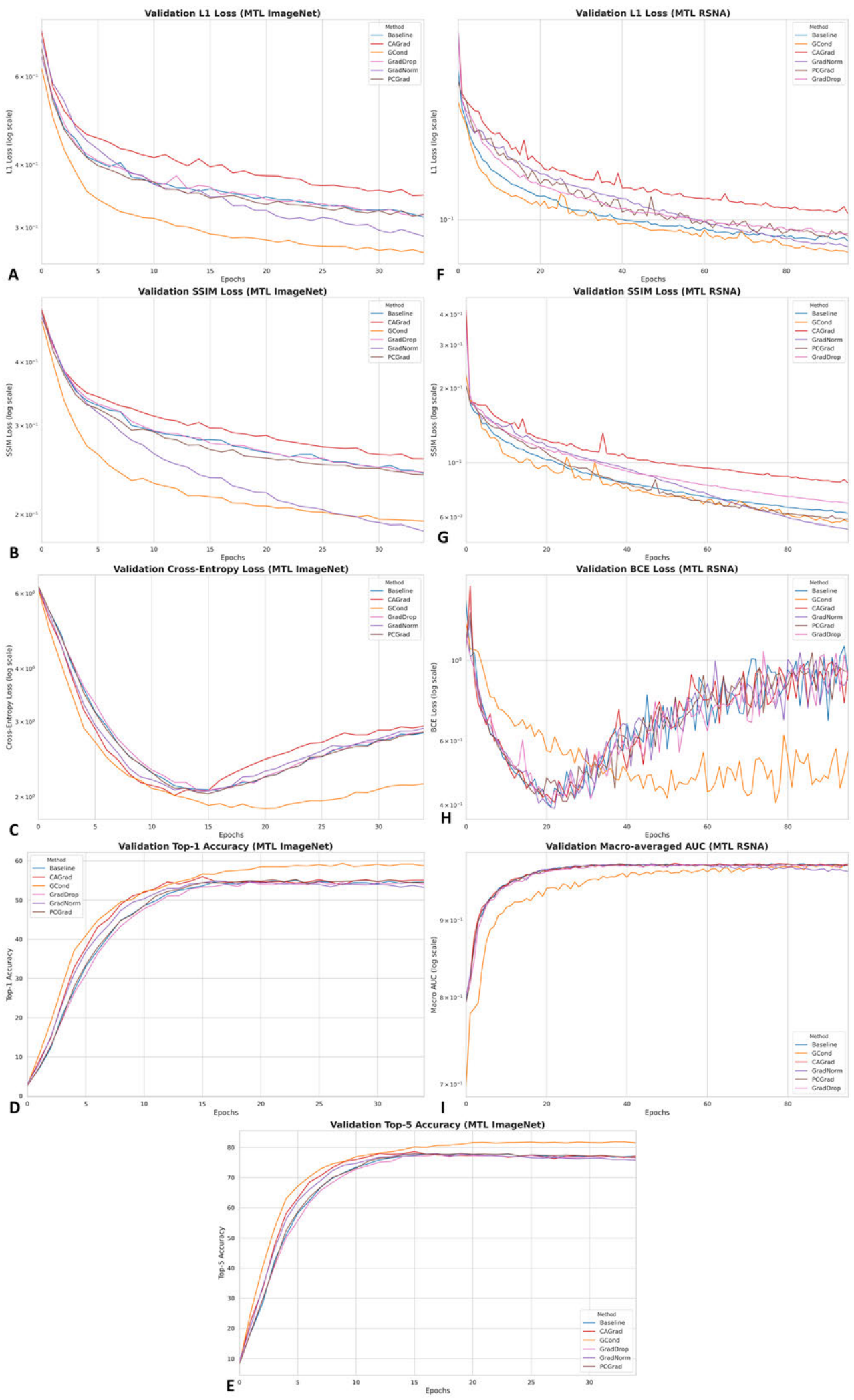

**FIGURE 13.** **Comparative validation dynamics of multi-task learning strategies on large-scale ImageNet-1K and RSNA CT dataset.**
**Note. Left Column (ImageNet-1k): (A) L1 Loss, (B) SSIM Loss, (C) Cross-Entropy Loss, (D) Top-1 Accuracy, and (E) Top-5 Accuracy. Right Column (RSNA CT): (F) L1 Loss, (G) SSIM Loss, (H) Binary Cross-Entropy (BCE)Loss, and (I) Macro AUC.**

### *C. ANALYSIS AND COMPARISON WITH PCGRAD*

Given that GCond's design was inspired by PCGrad, a direct comparison was performed between the original PCGrad, a "GCond-as-PCGrad" ablation, and the standard stochastic GCond. The proposed GCond module forms a unified gradient based on a global search for the most conflicting task pair, arbitration with a hybrid criterion (combining cosine stability to the previous step with relative norm strength), and a smooth projection with non-linear angle remapping, followed by EMA-smoothing with bias correction. As shown in (Fig. 8), both GCond versions exhibit significantly less variance and an absence of sharp spikes, indicating a more stable update direction. In terms of minimum cosine similarity, the original PCGrad remains in a strong conflict mode (values near -1), whereas GCond quickly moves the loss pair out of the critical zone thanks to its arbitration and modular projection.

The L1 and SSIM loss curves of the methods coincide only in the early phase, after which PCGrad gets stuck in a "tug-of-war" with pronounced oscillations, while GCond continues its monotonic descent. The quality and efficiency of the "GCond-as-PCGrad" conductor are noticeably lower than the stochastic mode, although they remain superior to the naive PCGrad variant with random projections due to the global conflict assessment and symmetric projection. The full GCond mode, with its smooth function and EMA-smoothing, further reduces noise and establishes a more confident direction. As a result, the optimizer takes larger, more reliable steps, which resolves the inter-task conflict and leads to accelerated convergence for both loss functions.

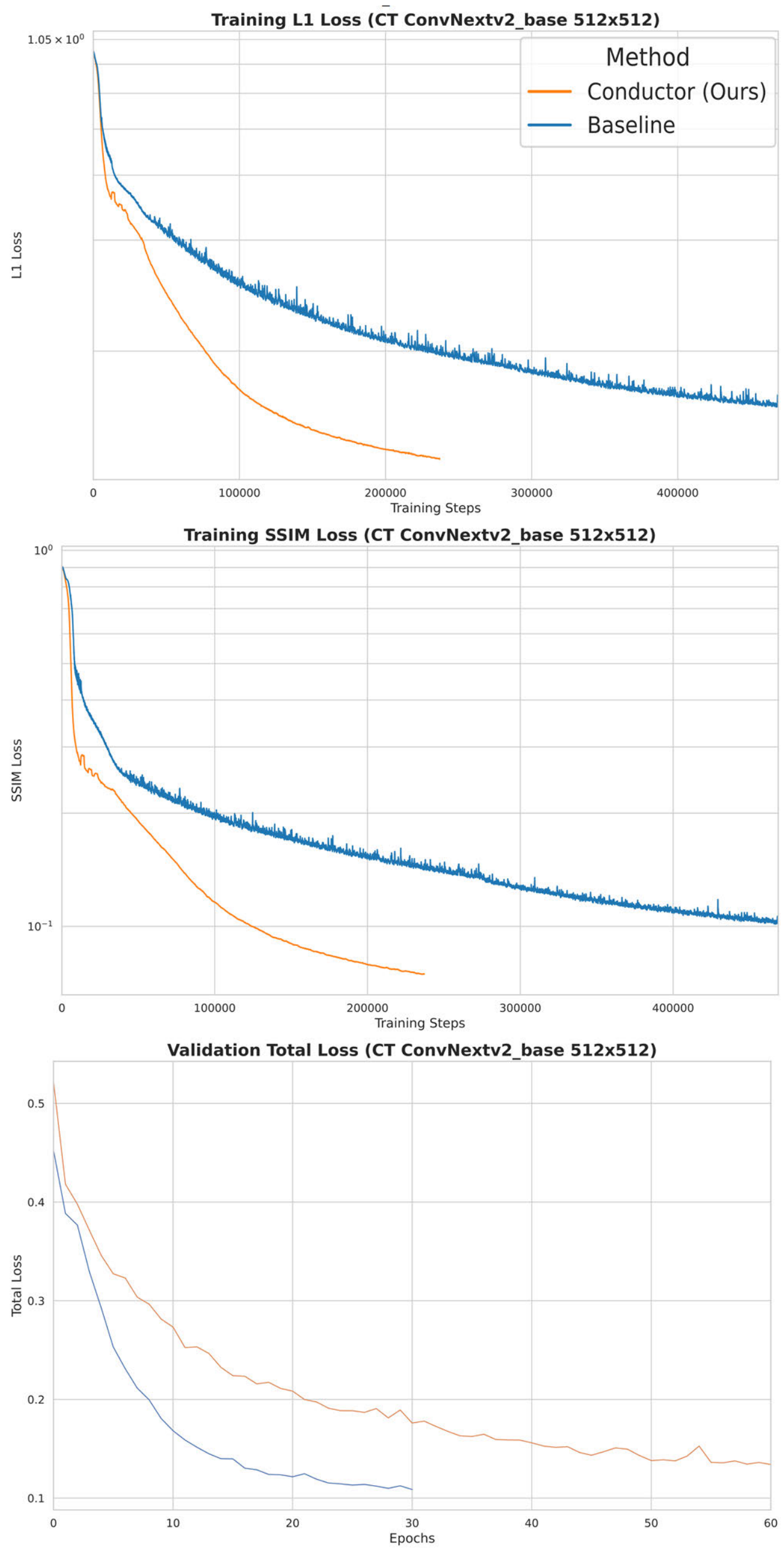

**FIGURE 14.** **Comparison of training quality between GCond and Baseline on the CT dataset using the ConvNeXtV2-Base mode.**

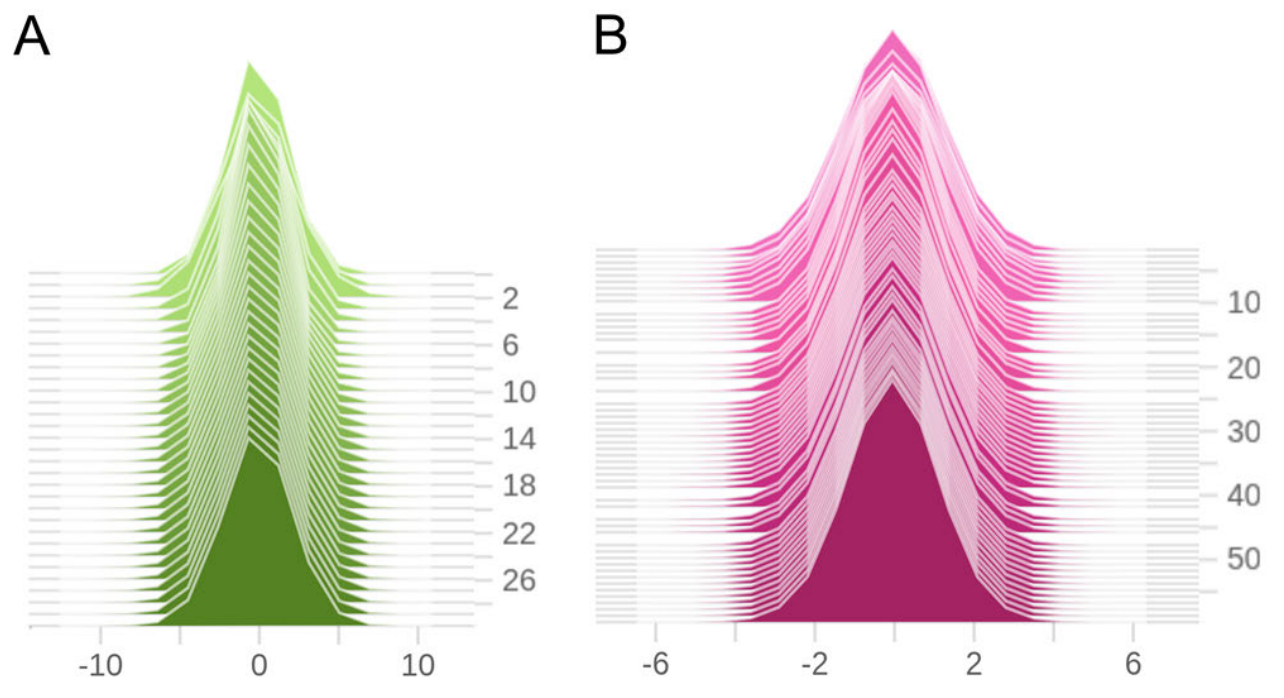

**FIGURE 15.** **Activation levels of the final encoder layer of ConvNeXtV2-Base by epoch.**
**Note. (A) GCond is on the left. (B) The linear combination(Baseline) is on the right.**

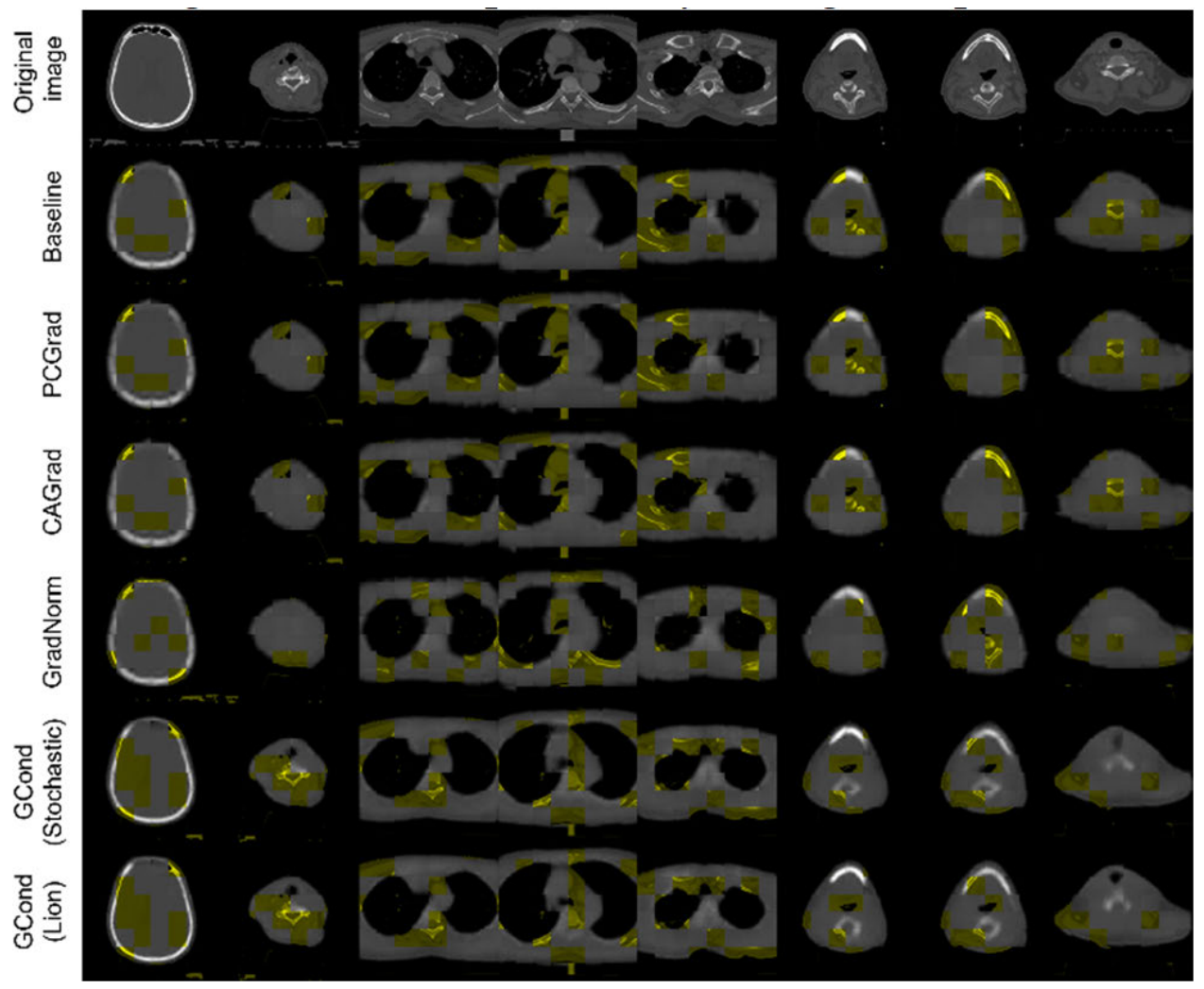

**FIGURE 16.** **Comparison of image reconstructions. The cells highlighted in yellow were visible to the model.**

## D. CURRENT LIMITATIONS

Several limitations should be considered when applying the proposed approach.

### 1) HYPERPARAMETER SENSITIVITY

GCond introduces hyperparameters-cosine similarity thresholds ($-0.8$, $-0.5$, $0.0$) for its tiered conflict resolution strategy, along with weights for the arbitration criteria. Although these defaults proved robust across two different datasets, a systematic analysis of their sensitivity and the potential development of auto-tuning mechanisms are required.

### 2) DEPENDENCE ON LARGE BATCH REGIMES

GCond is explicitly designed for regimes utilizing deep gradient accumulation to ensure low-variance gradient estimates. Consequently, the method's behavior in small-batch settings characterized by high-frequency gradient noise and high variance remains unexplored in this study. Our focus remains on scalability for large architectures where large effective batch sizes are inherent to the training process; thus, the application of GCond to small-batch regimes falls outside the scope of this work. It should also be emphasized that GCond is not a universal solution. In cases of persistent, nearly anti-parallel gradients (approaching 180°), the rigid "winner-takesall" scheme could potentially impair convergence. In practice, it may be advisable to first align task weights using simpler loss functions before applying GCond with a large effective batch.

**TABLE 7. Hyperparameters for ablation studies and baseline comparisons with column borders.**

| Parameter Group | Parameter | Baseline | PCGrad | CAGrad | GradNorm | Conductor (Pure) | Conductor (Stochastic) | Conductor (Sequential) | Conductor (Lion) |
|---|---|---|---|---|---|---|---|---|---|
| General | Optimizer | AdamW | | | | | | | Lion/Lars[1] |
| | Learning Rate | 2.0e-4 | | | | | | | 1.0e-5 |
| | Weight Decay | 0.05 | | | | | | | N/A |
| | Adam $\beta_1$ / $\beta_2$ | 0.9/0.95 | | | | | 0.0/0.95 | | N/A |
| | $L_1$ / SSIM $\lambda$ | 0.85/0.15 | | | Dynamic | 0.85/0.15 | | | |
| Strategy-Specific | GradNorm $\alpha$ | N/A | | | 1.5 | N/A | | | |
| | GradNorm $\lambda$ LR | N/A | | | 1.0e-3 | N/A | | | |
| Conductor: Core | return_raw_grad | N/A | | | | True | False | False | False |
| | stochastic_accumulation | N/A | | | | True | True | False | True |
| | use_smooth_logic | N/A | | | | True | | | |
| | remap_power | N/A | | | | 2.0 | | | |
| Conductor: Optimizer | use_lion | N/A | | | | False | False | False | True |
| | momentum_beta ($\beta_1$) | N/A | | | | 0.0 | 0.9 | 0.9 | 0.9 |
| | trust_ratio_coef (LR) | N/A | | | | N/A | N/A | N/A | LR Sched.[2] |
| | trust_ratio_clip | N/A | | | | N/A | N/A | N/A | 50.0 |
| Conductor: Projection | projection_max_iters | N/A | | | | 3 | | | |
| | norm_cap | N/A | | | | None | | | |
| | conflict_thresholds | N/A | | | | -0.8,-0.5,0 | | | |
| Conductor: Arbitrator | dominance_window | N/A | | | | 0 - disabled | | | |
| | norm_ema_beta | N/A | | | | 0.95 | | | |
| | tie_breaking_weights | N/A | | | | (0.8, 0.2) | | | |

**TABLE 7.** *(Continued).* **Hyperparameters for ablation studies and baseline comparisons with column borders.**

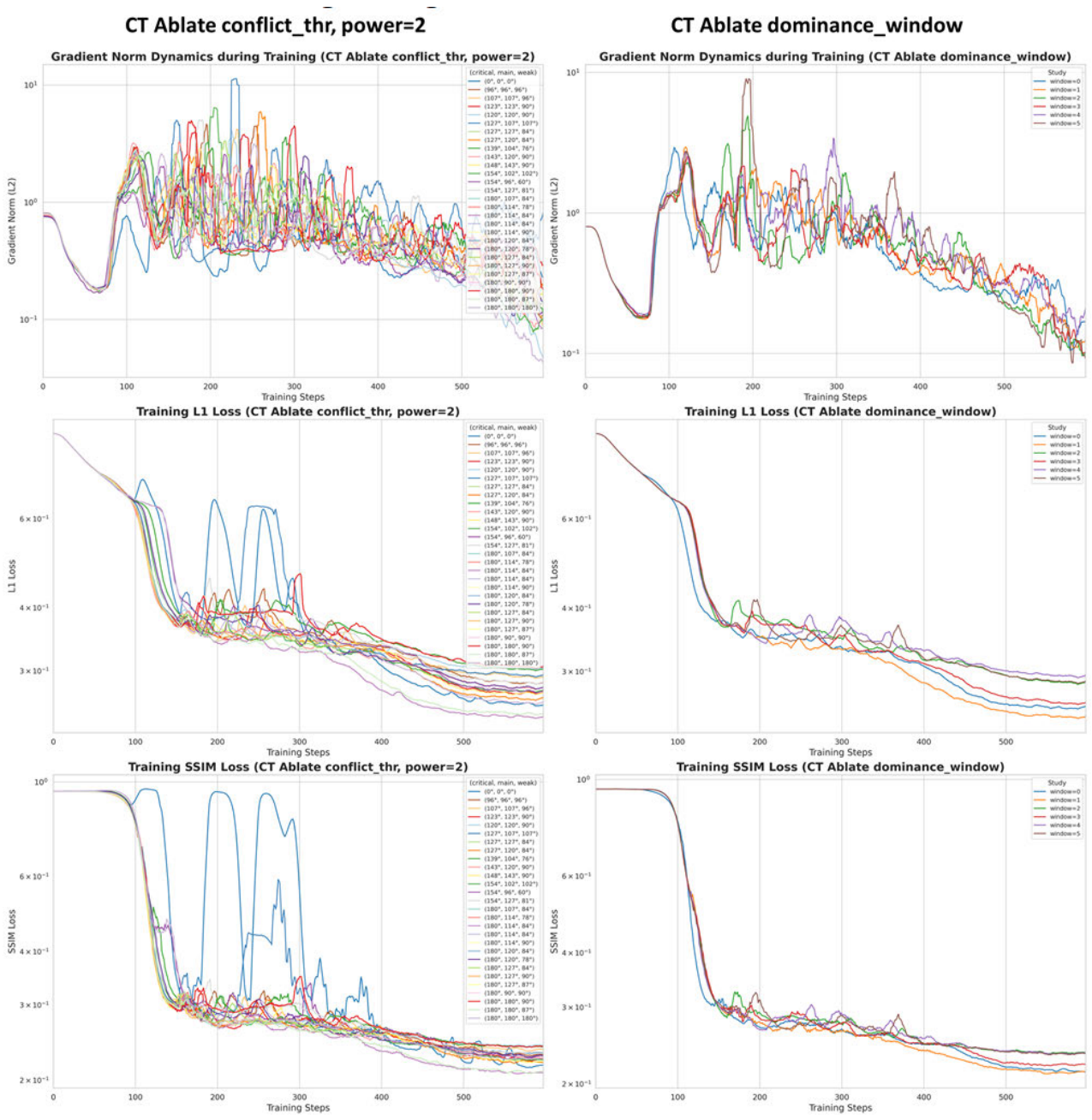

**FIGURE 17. Ablation study of geometric thresholds and dominance window.**
**Note. x-axis: (Left) Training dynamics (Gradient Norm, L1 Loss, and SSIM Loss) across various conflict_thresholds configurations using smooth logic (power=2). (Right) Impact of the dominance_window (w) on convergence. While w=3 demonstrates optimal performance, w=0 is maintained as the baseline to ensure a stateless comparison of gradient projection properties.**

### 3) FRAMEWORK DEPENDENCIES

The current implementation of GCond is tailored for PyTorch $\geq$ 2.0 and utilizes torch.func.functionalcall, AMP/GradScaler, and optional DDP synchronization. Porting it to other frameworks or versions $<$ 2.0 would require adapting these components.

## *E. PRACTICAL IMPLICATIONS*

GCond presents a scalable solution for multi-task learning that can be easily integrated into existing PyTorch $\geq$ 2.0-based deep learning frameworks without significant architectural changes. It can be incorporated as a "drop-in" component, as it operates on top of an existing gradient accumulation loop and writes the unified gradient vector directly to p.grad. This integration requires minimal code modifications and remains compatible with any optimizer and GradScaler (it is recommended to calculate momentum within the module). Moreover, GCond enables the formation of larger effective batches and the use of deeper architectures under the same hardware constraints, offering more efficient utilization of computational resources. Thus, the proposed GCond approach represents a practical solution for effectively resolving gradient conflicts in modern, large-scale multi-task learning models, opening new avenues for research in this field.

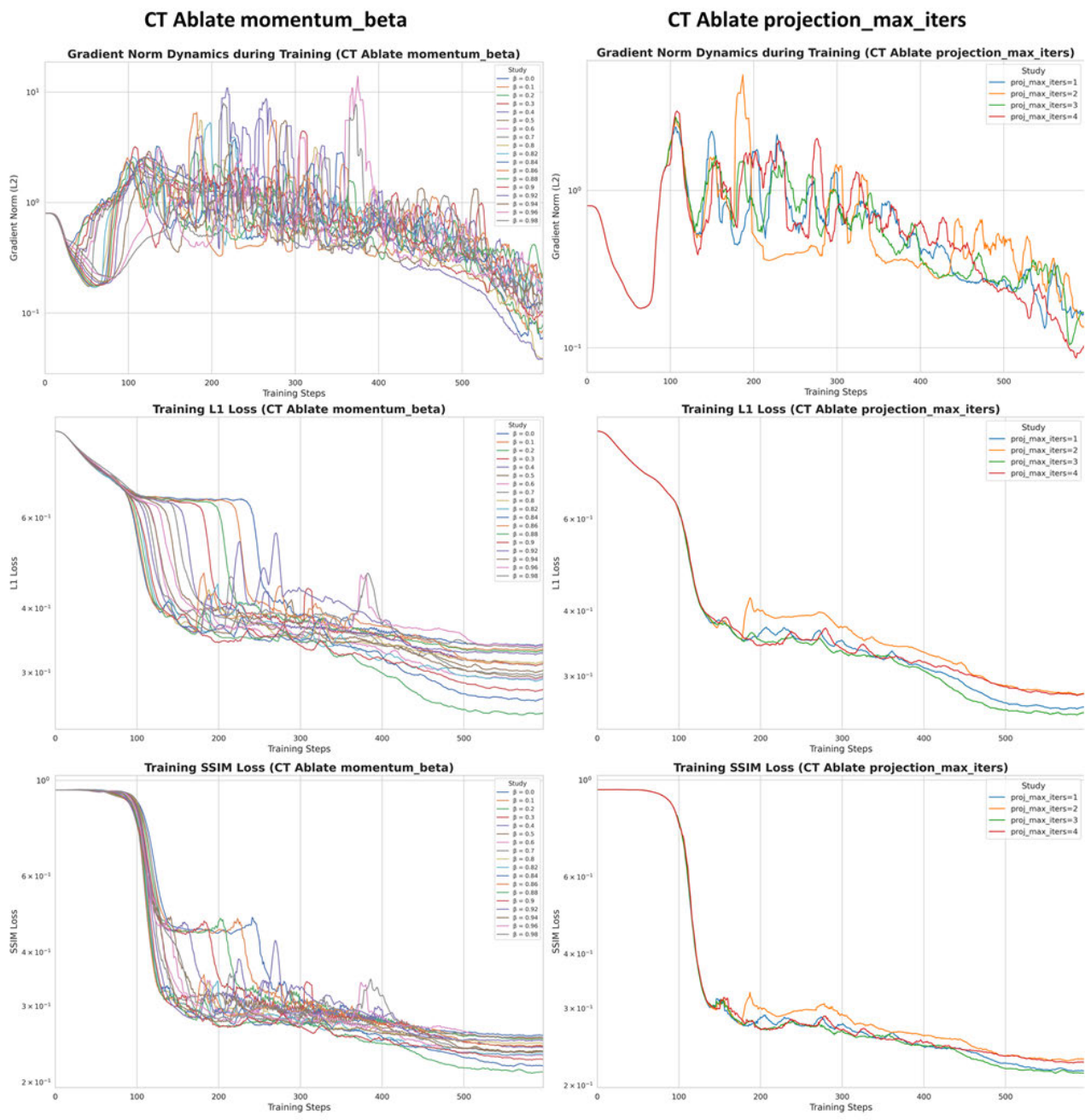

**FIGURE 18.** **Ablation study of momentum and projection hyperparameters.**
**Note. Left: Influence of the momentum_beta parameter on training stability, illustrating the role of higher values(e.g., 0.9) in mitigating stochastic noise. Right: Effect of the projection_max_iters on convergence rates, comparing iterations from 1 to 4.**

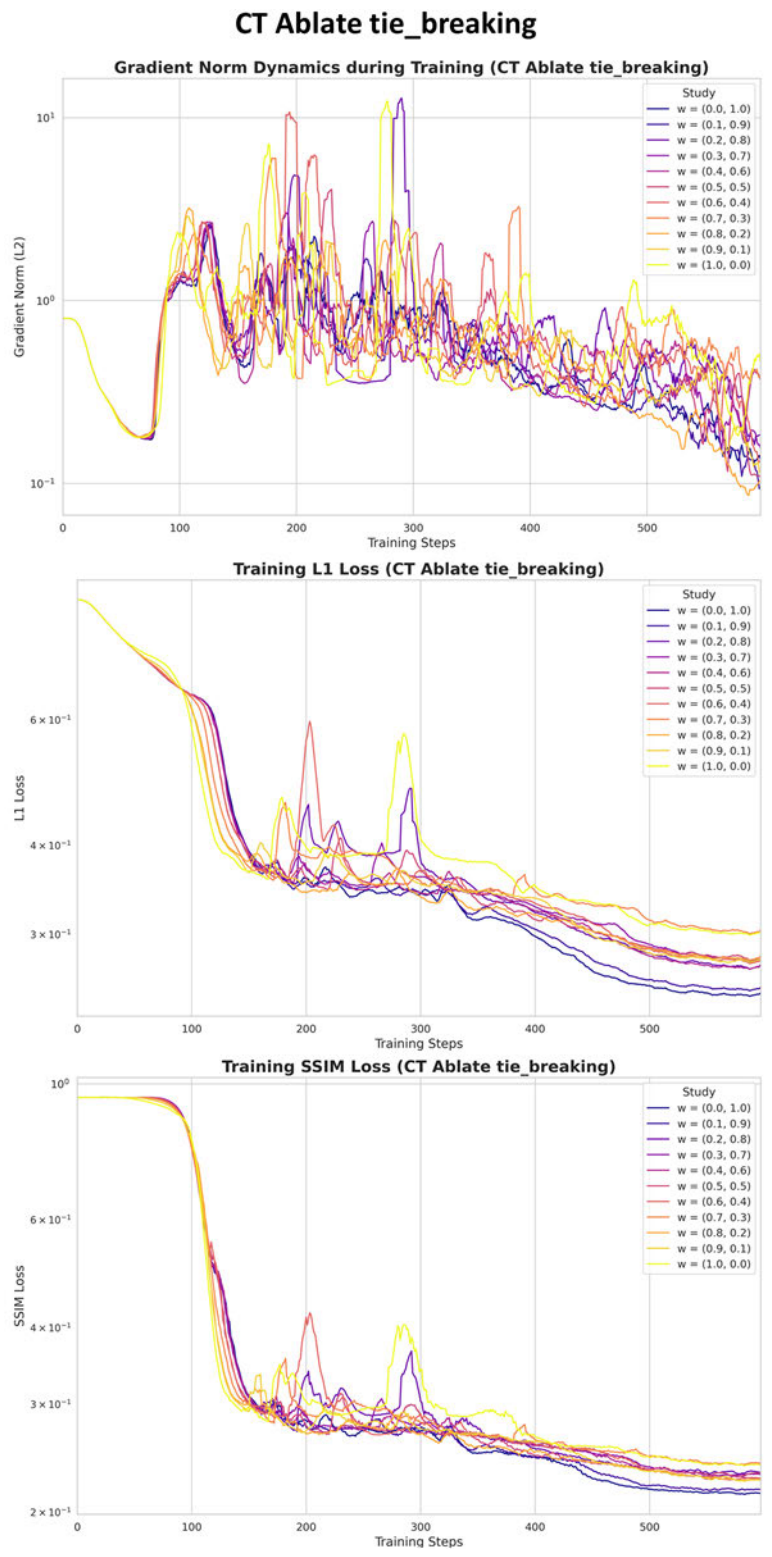

**FIGURE 19.** **Ablation study the asymmetric tie-breaking arbitration policy. Note. x-axis: Training steps, y-axis (top row): Gradient Norm, y-axis (middle row): L1 Loss, y-axis(bottom row): SSIM Loss.**

## V. CONCLUSION

We have introduced Gradient Conductor (GCond), a novel approach to multi-task learning that addresses the fundamental limitations of existing gradient surgery methods. Through an ''accumulate-then-resolve'' paradigm and adaptive arbitration, GCond achieves superior performance while maintaining excellent scalability properties. Our experiments demonstrate significant improvements over SOTA on both medical imaging and natural image datasets. The superiority of GCond stems from its core mechanism: a synergy of gradient accumulation and adaptive arbitration. This allows it to form a statistically reliable gradient direction by analyzing the optimization history and suppressing the noise of stochastic estimates. As a result, the final smoothed gradient enables modern optimizers like AdamW or Lion/LARS to take more confident steps, which translates directly into superior performance in both quantitative metrics and computational efficiency. A key feature is the ability to operate in a stochastic mode, providing an N-fold (where N is the number of loss functions) performance increase while maintaining optimization quality. Another advantage is its scalable architecture, which allows it to work with large models like ConvNeXtV2-Base where traditional methods in their original implementations are inapplicable. The high robustness of its hyperparameters and its successful integration with adaptive optimizers make GCond suitable for a wide class of deep learning tasks. The application of GCond has the potential to make a significant contribution to various fields of digital pathology, including complex histological image analysis. This work opens up prospects for further research into the connection between conflict-resilient optimization and the generalization capabilities of neural networks. However, conclusions regarding the superiority of GCond must be framed with appropriate caution. This study explicitly clarifies that the observed advantages are limited to the chosen large-batch computational regime enabled by gradient accumulation. The performance of GCond in traditional small-batch, high-variance MTL settings such as those typical of NYUv2 and Cityscapes benchmarks remains unexplored and falls outside the scope of this research.

## APPENDIX

### A. SCHEMATIC REPRESENTATION OF THE GCOND'S WORK

The algorithm (Fig. 9) determines the projection strategy based on the cosine similarity (C) between conflicting gradient pairs. Check Zones: The decision logic classifies interaction into four categories: Agreement, Mild Conflict, Moderate Conflict, and Critical Conflict. For Moderate and Critical zones, a "winner" is selected using a hybrid score. Based on the zone and winner selection, the method applies either symmetric projection, PCGrad-style orthogonal projection, or a "Winner-Takes-All" strategy (zeroing out the loser's gradient).

### B. VALIDATION

The validation curves (Fig. 10), demonstrate that the differences between the learning curves stochastic and the exact (sequential) modes GCond are statistically insignificant.

The trajectories of the stochastic GCond mode are consistently below the curves of the SOTA methods (Fig. 11), from the early epochs to the final plateau. This gap is maintained with less variance between epochs, confirming the proposed approach's ability to provide a more stable, conflict-resilient estimate of the optimization step.

The figure 13 display validation metrics across training epochs for the ConvNeXtV2-Base architecture. On the large-scale ImageNet task, GCond (orange curve) acts as a plasticity facilitator, eliminating the U-shaped overfitting curve in Cross-Entropy loss observed in baselines and achieving a monotonic increase in Top-1 Accuracy (+4.2% vs. Baseline). Conversely, on the smaller RSNA dataset, GCond acts as a regularizer, preventing confidence overfitting where the BCE of competing methods diverges (>1.0) despite high AUC scores. Across both domains, GCond consistently yields the lowest reconstruction errors (L1/SSIM).

### C. ADDITIONALL EXPERIMENTS WITH THE CONVNEXTV2-BASE ARCHITECTURE

Final testing on the complex ConvNeXtV2-Base model was performed on a system with an NVIDIA H200 GPU (140 GB VRAM) and a 48-core Intel(R) XEON ®PLATINUM 8568Y+ CPU. Its software environment included PyTorch 2.5.1, Python 3.10.18, Ubuntu 22.04, and CUDA 12.4.

Additionally, experiments with the ConvNeXtV2-Base architecture on SSL task with L1 and SSIM losses on the CT HN dataset with 512 × 512 resolution images demonstrated significant improvements in convergence and stability. With the GCond model, the L1 and SSIM training curves were characterized by substantially greater smoothness and lower variance compared to the Baseline. On the validation set, by the 15th epoch (10 of which were a warmup period), GCond had already reached and subsequently surpassed the final performance level of the Baseline model trained for 60 epochs. When training was stopped at the 30th epoch, a noticeable gap in both loss functions persisted Fig. 14. We intentionally limited GCond's training to 30 epochs, as the goal of this experiment was to demonstrate the method's potential efficiency-that is, GCond achieves lower loss values in the same or fewer epochs.

Monitoring the activations of the final encoder layer Fig. 15 revealed a nearly twofold expansion in the distribution of activations from the very first epochs. We interpret this as an indicator of a richer, noncollapsing representation and a consequence of the high signal-to-noise ratio in the gradient and the conflict-resilient projection in GCond, which enhances both optimization speed and the model's potential generalization capability. A detailed analysis of the large model's representations is a subject for our future work.

### D. COMPARISON OF IMAGE RECONSTRUCTIONS

We further investigated whether the failure of standard projection methods, or the gains from our smoothing, could be replicated by simply increasing the temporal averaging of the AdamW optimizer. An ablation study was conducted on the optimizer's hyperparameters, despite the use of a large effective batch size (6144) which already provides substantial stabilization. Modifying the first-moment parameter ($\beta_1$) to 0.99 yielded no improvements, confirming its established optimality near 0.9. More critically, tuning the second-moment parameter ($\beta_2$) to high-stability values (0.999 and 0.9999) also resulted in marginally worse performance than the standard 0.95 default. This strongly suggests that the superior stability GCond achieves stems from its explicit, adaptive arbitration and internal momentum (as shown in Fig. 16), rather than from simple temporal smoothing that can be replicated by tuning the optimizer.

It is important to note that the SOTA methods are not capable of reconstructing CT images Fig. 16, whereas the GCond's approaches contour bones and other more complex structures such as soft tissues.

### E. ARCHITECTURE PARAMETERS

#### 1) CONSTANT PARAMETERS

The following parameters were held constant across all training runs described in the main table below:

1) Model Architecture:
   - Encoder: mobilenetv3 small 100
   - Decoder Embedding Dimension: 256
   - Decoder Depth (Transformer Blocks): 2
   - MAE Mask Ratio: 0.75
- Data & Augmentation:
   - Image Size: 256 × 256
   - Patch Size: 32 × 32
   - CT Dataset Mean / Std: 0.107961 / 0.173622
   - ImageNet Dataset Mean / Std: [0.485, 0.456, 0.406] / [0.229, 0.224, 0.225]
- Training Framework:
   - Epochs: 15
   - Total Batch Size: 256
   - Accumulation Steps: 24
   - Workers: 18
   - Gradient Clipping Norm: 50.0
- LR Scheduler:
   - Type: Linear Warmup + Cosine Annealing

- Warmup Epochs: 2
- Eta Minimum (Cosine): 1e-6
- Large-scale ConvNeXtV2-Base Constant Parameters:
- General Architecture & Optimization:
- Model: ConvNeXtV2-Base
- Precision: bfloat16 mixed precision
- Optimizer: AdamW (Weight Decay: 0.05)
- Momentum Strategy (GCond): External $\beta$1=0 / Internal $\beta$ =0.9
- Warmup Scheduler: Linear (3 epochs)
- Early Stopping: Disabled
- RSNA CT Configuration:
- Input Resolution: 416 × 416
- Effective Batch Size: 1,980
- Learning Rate: 1.5 × 10−4
- Training Duration: 95 Epochs
- Loss Weights: λcls=0.1, λL1=0.7, λSSIM=0.2
- ImageNet-1k Configuration:
- Input Resolution: 256 × 256
- Effective Batch Size: 2,952
- Learning Rate: 2.0 × 10−4
- Training Duration: 35 Epochs
- Loss Weights: λcls=0.15, λL1=0.65, λSSIM=0.2

### F. HYPERPARAMETERS

To ensure the reproducibility of our results, this section details the complete set of hyperparameters used for each experimental condition. All experiments were conducted using three different random seeds (11, 42, 2025) to report mean performance and confidence intervals.

#### 1) ABLATION STUDY AND BASELINE PARAMETERS

The following table outlines the parameters that were varied across the different gradient management strategies. "N/A" indicates that a parameter is not applicable to the specified strategy.

### G. HYPERPARAMETER ANALYSIS

A comprehensive ablation study of key hyperparameters was conducted to determine the optimal configuration for the GCond module. A systematic grid search on the CT image dataset with the combined L1 and SSIM loss function identified the most robust configuration, providing an effective balance between stability and convergence speed. Specifically, the optimal arbitration weights were tie breaking weights = (0.8, 0.2), which prioritize gradient direction stability over its instan taneous magnitude. The optimal three-zone conflict threshold system was conflict thresholds = (-0.8, -0.5, 0), which clearly delineates the intervention modes.

#### 1) THRESHOLD VALUES

The threshold configuration of (-0.8, -0.5, 0.0) demonstrated the best long-term convergence. A value of $\theta$weak = 0.0 serves as an intuitive boundary between conflict and agreement. The lower threshold $\theta_{crit} = -0.8$ activates asymmetric conflict resolution only when vectors are significantly anti-parallel, preventing overly aggressive correction for moderate disagreements. Notably, in the early training stages (epochs 2-5), the configuration $\theta = (-1.0, -0.4, 0.1)$ yielded faster results. Here, $\theta_{weak} = 0.1$ treated even slight agreement as a conflict, subjecting the gradients to symmetric orthogonalization. This accelerated the initial direction finding but later restricted exploration of the parameter space, leading to stagnation in a local optimum.

#### 2) ARBITRATION WEIGHTS

For tie breaking weights, an imbalance proved preferable to equilibrium. It allows the arbiter to consistently adhere to one strategy (gradient stability or relative strength), avoiding oscillations in decision-making when resolving critical conflicts.

#### 3) MOMENTUM

The momentum beta parameter showed optimal results in the range around 0.9, consistent with common practices for adaptive optimizers.

#### 4) DOMINANCE WINDOW

Our ablation study on the dominance window (w) hyperparameter revealed that w=3 and w=0 (disabled) yielded the top results. However, the dominance mechanism is stateful, making the gradient update dependent on the optimization history. To ensure a fair and stateless comparison across methods, thereby isolating their intrinsic properties, we set w=0 for all primary experiments. While this guarantees reproducibility and a controlled evaluation, we acknowledge that a non-zero window could be optimal in practical applications to prevent task starvation. Despite the presence of numerous hyperparameters, our analysis showed that the proposed set of default values, initially selected for medical segmentation tasks, demonstrates a high degree of versatility and ensures stable convergence across diverse datasets, including ImageNet.

## ACKNOWLEDGMENT

The authors wish to thank The Cancer Imaging Archive (TCIA) and the Radiological Society of North America (RSNA) for providing the public datasets used in this study. They also acknowledge the use of Google's Gemini for assistance with the translation of the manuscript from Russian to English and for improving language and grammar. They reviewed and edited the text and bear full responsibility for the final content.

### ETHICAL STATEMENT

This study was conducted in strict accordance with the ethical principles outlined in the Declaration of Helsinki. The research utilizes exclusively retrospective, de-identified data retrieved from publicly available repositories. After gaining access to the datasets, non-disclosure agreements were signed to prevent the data from being shared with unauthorized parties. The data was only used for conducting experiments. As the study involves the analysis of secondary data without any direct interaction with human subjects or access to identifiable private information, it is exempt from Institutional Review Board approval and informed consent requirements according to standard academic and institutional guidelines. Throughout the study, strict adherence to academic ethical principles was maintained.

### DATA AVAILABILITY

The following information was supplied regarding data availability: The source code and experimental data for this study are available in a Zenodo repository (DOI: 10.5281/zenodo.18514611) and on GitHub (https://github.com/AlevLab-dev/GCond). The datasets are available at:

1. RSNA 2022 Cervical Spine Fracture Detection: https://www.kaggle.com/competitions/rsna-2022-cervical-spine-fracture-detection/
2. RSNA Intracranial Hemorrhage Detection: https://www.kaggle.com/competitions/rsna-intracranial-hemorrhage-detection
3. RADCURE from The Cancer Imaging Archive: https://www.cancerimagingarchive.net/collection/radcure/
4. Benjamin paine Imagenet-1k-256 × 256: https://huggingface.co/datasets/benjamin-paine/imagenet-1k-256 × 256

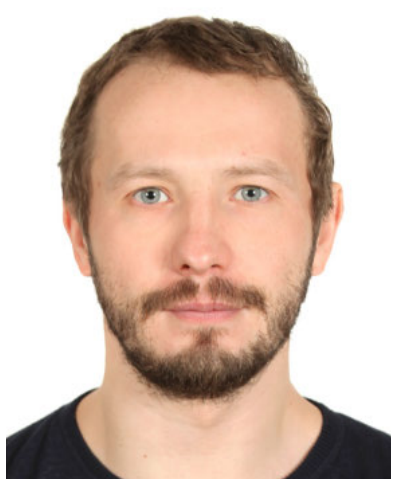

**EVGENY ALVES LIMARENKO** received the specialist degree in medical physics from Moscow Engineering Physics Institute (MEPhI), in 2007. He is currently pursuing the master's degree in applied mathematics and physics with Moscow Institute of Physics and Technology (MIPT). With more than 20 years of experience in software architecture and C++ development, his research interests include the intersection of computer science and medicine, specifically in deep learning, computer vision, and self-supervised learning for medical image analysis.

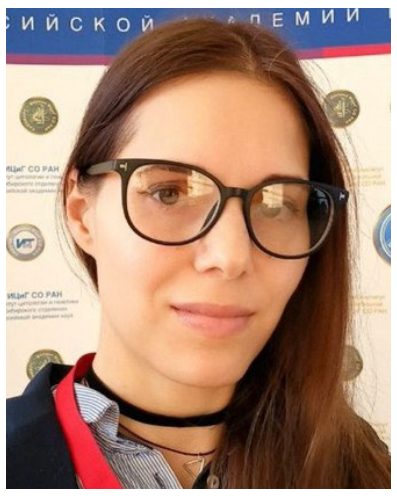

**ANASTASIIA STUDENIKINA** graduated from Novosibirsk State Medical University, Novosibirsk, Russia in 2017, and the Ph.D. degree in 2021. She is currently pursuing the master's degree in applied mathematics and physics with the Moscow Institute of Physics and Technology (MIPT), Russia. Her research interests include molecular mechanisms of cancer development, cancer biomarker identification, and deep learning algorithms for analyzing tumors and their microenvironment.

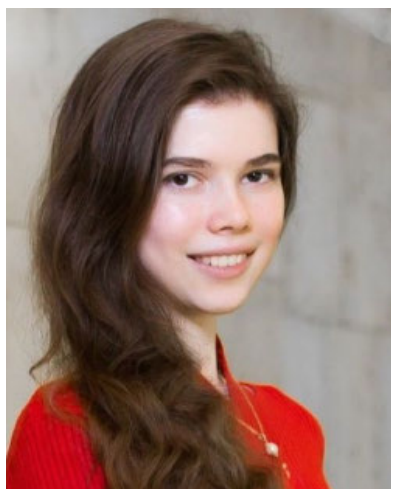

**SVETLANA ILLARIONOVA** received the Ph.D. degree in computer science from Skoltech in 2022. She is the Head of the Research Group in Computer Vision at Skolkovo Institute of Science and Technology (Skoltech), Moscow, Russia. Her research focuses on developing novel computer vision and generative AI methodologies for knowledge-intensive domains, including histopathology, remote sensing, and environmental monitoring. She specializes in multi-modal data fusion, explainable AI, and the application of deep learning to biomedical and earth observation data.

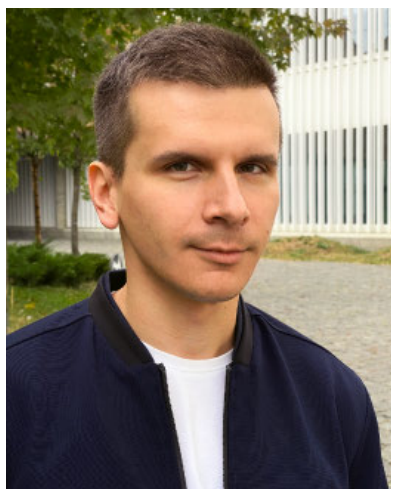

**MAXIM SHARAEV** received the Ph.D. degree in physics and mathematics from Lomonosov Moscow State University, Moscow, Russia. He is currently the Head of the Biomedically-informed AI Laboratory at the Artificial Intelligence Center, Skolkovo Institute of Science and Technology and the University of Sharjah, Sharjah, UAE. His research interests include multimodal data analysis, signal and image processing, neuroimaging, and the development of AI-based methods for biomedical applications.